\documentclass{article}
\usepackage[main, final]{neurips_2026}

\usepackage[utf8]{inputenc} 
\usepackage[T1]{fontenc}    
\usepackage{hyperref}       
\usepackage{url}            
\usepackage{booktabs}       
\usepackage{amsfonts}       
\usepackage{nicefrac}       
\usepackage{microtype}      
\usepackage{xcolor}         
\usepackage{svg}

\usepackage{graphicx}
\usepackage{amsmath}
\usepackage{multirow}
\usepackage{algorithm}
\usepackage{algorithmic}

\usepackage{longtable}
\usepackage{seqsplit}

\usepackage{amssymb}
\usepackage{array}
\newcolumntype{L}[1]{>{\raggedright\arraybackslash}p{#1}}
\newcommand{\rankone}[1]{\textcolor[HTML]{0072B2}{\textbf{#1}}}
\newcommand{\ranktwo}[1]{\textcolor[HTML]{E69F00}{\textbf{#1}}}
\newcommand{\rankthree}[1]{\textcolor[HTML]{009E73}{\textbf{#1}}}
\title{Amortized Molecular Optimization via Group Relative Policy Optimization}

\author{
  Muhammad bin Javaid\thanks{Equal contribution.}\textsuperscript{\, \, 1} \And
  Hasham Hussain\footnotemark[1]\textsuperscript{\, \, 1, 2} \And
  Ashima Khanna-Reiter\textsuperscript{3} \And
  Berke Ki\c{s}in\textsuperscript{1} \And
  Jonathan Pirnay\textsuperscript{3} \And
  Alexander Mitsos\textsuperscript{4} \And
  Dominik G. Grimm\textsuperscript{3, 5} \And
  Martin Grohe\textsuperscript{1} \\
  \\
  \textsuperscript{1}RWTH Aachen University, Department of Computer Science, Aachen, Germany \\
  \textsuperscript{2}Alfred E. Tiefenbacher (GmbH \& Co. KG), Hamburg, Germany \\
  \textsuperscript{3}Technical University of Munich, TUM Campus Straubing for Biotechnology and Sustainability \\
  \textsuperscript{4}RWTH Aachen University, Process Systems Engineering (AVT.SVT), Aachen, Germany \\
  \textsuperscript{5}Technical University of Munich, TUM School of Computation, Information and Technology \\
  \texttt{javaid@informatik.rwth-aachen.de}
}

\begin{document}

\maketitle

\begin{abstract}
    In structurally constrained molecular optimization, state-of-the-art methods
treat every new input structure as a separate instance, restarting an expensive
oracle-driven search from scratch and scaling poorly to settings with many
starting structures or expensive oracles. While amortized approaches that learn a
transferable policy could in principle remove this bottleneck, existing
methods struggle to generalize to diverse structural constraints at
inference time. We present AMORTIX, a Graph Transformer model that natively
supports such constraints, optimizing molecular structures in a single rollout
with zero inference-time oracle calls. A central challenge for amortized
training in this domain is that optimization difficulty varies drastically
across starting structures. We address this by centering rewards within groups
of completions that share the same starting structure. We evaluate on goal-directed scaffold decoration, including single- and
multi-target kinase inhibitor design, and on a few-shot prodrug case study.
AMORTIX outperforms both amortized baselines and search-based instance optimizers on
kinase scaffold decoration, and ablations across 23 objectives show that
group-relative centering improves over REINFORCE and PPO and is substantially
more consistent across scaffold splits. The prodrug case study further
demonstrates transfer of learned modification rules to unseen drug structures.
Our code is available at \url{https://github.com/mbinjavaid/AMORTIX}.
\end{abstract}

\section{Introduction}\label{sec:intro}

\begin{figure*}[t]
    \centering
    \includegraphics[scale=0.64]{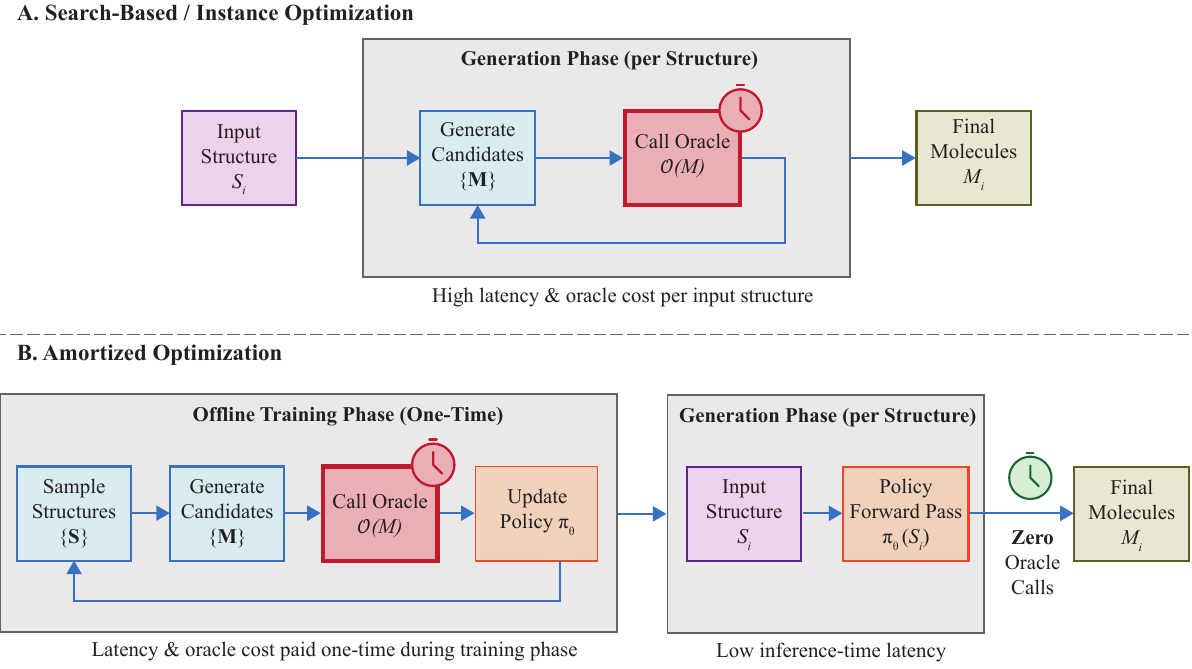}
    \caption{Comparison of optimization paradigms. \textbf{(A) Instance Optimization:} Requires an expensive iterative search with hundreds to thousands of oracle calls for every new input structure $S_i$, resulting in high cost that scales linearly with library size. \textbf{(B) Amortized Optimization:} Front-loads computation into offline training. The learned policy $\pi_\theta$ generates optimized molecules for new, unseen inputs $S_i$ in a single rollout without inference-time oracle calls, enabling scalable, high-throughput design.}
    \label{fig:instance_vs_amortized}
\end{figure*}

Molecular design for drug discovery often requires modifying or elaborating a given starting structure rather than designing from scratch. Tasks such as scaffold decoration, linker design, and lead optimization all begin from a fixed molecular substructure that must be preserved, and are scored against oracles whose cost ranges from milliseconds for heuristic proxies to hours for high-fidelity simulations such as free-energy perturbation. Ideally,
a generative model would propose optimized modifications in a single rollout, amortizing the search cost over the training period (Figure~\ref{fig:instance_vs_amortized}B).

In existing approaches, the dominant paradigm is the opposite: \textbf{instance
optimization}. Genetic algorithms~\cite{mol_GA, genetic_gfn} and guided discrete diffusion models~\cite{genmol} achieve strong results by issuing hundreds to thousands of oracle calls per design task, restarting the search from scratch for every new input structure (Figure~\ref{fig:instance_vs_amortized}A). The PMO benchmark~\cite{pmo} captures this cost on isolated tasks through a strict
$10{,}000$-call budget, but does not reflect the cumulative cost across many starting structures or the per-call cost of high-fidelity oracles. An \textbf{amortized} conditional policy $\pi_\theta(y \mid x)$, trained once, could generate optimized modifications $y$ for any unseen starting structure $x$ in a single rollout with no inference-time oracle calls. In practice, however, few amortized approaches natively support hard structural
constraints, and those that do have not matched instance optimization on optimization tasks.
We trace this failure to a specific cause: the difficulty of optimization varies dramatically across starting structures. Some structures yield high-scoring derivatives easily, while others occupy chemically constrained regions where even the best modifications score modestly. A global baseline
cannot distinguish a strong policy decision on a hard structure from a trivial modification on an easy one; the result is high-variance advantage estimates dominated by the easy instances.

We address this with AMORTIX, a Graph Transformer-based policy trained with reinforcement learning (RL) for amortized molecular design. The policy constructs molecules step by step, adding atoms and bonds under a valence-based action mask that enforces chemical validity by construction. For training, we adapt Group Relative Policy Optimization (GRPO)~\cite{grpo} to the molecular domain: rather than estimating a global baseline, we sample a group of $G$ completed trajectories from each
starting structure and center rewards relative to the group mean. Each starting structure's completions are thus scored only against peers also starting from the same structure, decoupling the learning signal from cross-structure difficulty variation.

Our contributions are as follows. We identify heterogeneous optimization
difficulty across starting structures as a primary failure mode for amortized RL
in structurally-constrained molecular optimization, and adapt GRPO to address
it. We show empirically that standard approaches do not resolve this: both
REINFORCE with a global baseline and PPO with a learned value critic reach lower
scores on most objectives and are considerably less consistent across scaffold
splits than our approach (Section~\ref{subsec:ablation}). Our proposed method, AMORTIX, generalizes to held-out scaffolds with zero inference-time oracle calls, and outperforms both amortized
and instance-optimization baselines on structurally-constrained kinase optimization tasks. In terms of oracle calls, its one-time training cost is repaid after a modest number of test scaffolds,
with the savings growing from there (Figure~\ref{fig:oracle_cost}). Finally, a few-shot prodrug case study shows transfer of learned modification rules to 52 unseen parent drugs. As a \textit{de novo} sanity check
of the policy architecture, AMORTIX is also competitive on the PMO benchmark
(Appendix~\ref{pmo}).

\section{Related Work}
\begin{figure*}[t]
    \centering
    \includegraphics[scale=0.99]{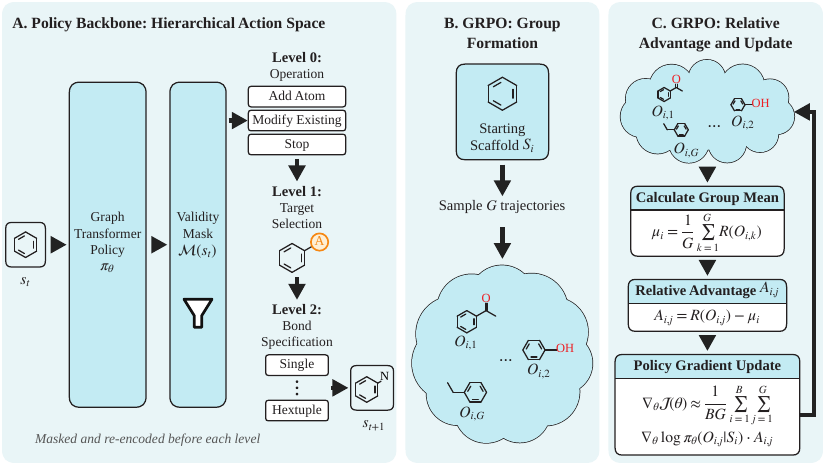}
    \caption{Overview of AMORTIX. \textbf{(A)} The policy decodes a composite action
in three levels (operation, target atom, bond order) under a validity mask.
\textbf{(B)} For each starting scaffold, $G$ completions are sampled
independently from the policy. \textbf{(C)} Rewards are centered by the group
mean, and the resulting advantages weight the policy-gradient update.}
\label{fig:grxfrom}
\end{figure*}

Generative molecular design methods can be broadly classified into two primary paradigms: instance optimization and amortized optimization. While significant progress has been made in both areas, the specific regime of amortized optimization under strict structural constraints is comparatively underserved. Existing methods in this space either rely on expensive search-based imitation, struggle to scale across diverse structures, or use global rewards that fail when structural difficulty varies heterogeneously. AMORTIX specifically targets this gap. For a comprehensive taxonomy illustrating where AMORTIX sits relative to existing approaches, see Appendix~\ref{appendixA} and Figure~\ref{fig:taxonomy}.

\subsection{Instance Optimization}
While amortized policies evaluate external scoring functions primarily during training, methods in this category actively query these oracles to guide an iterative traversal of chemical space at inference time. Because they treat each generation task as a distinct search problem to be solved from scratch, this paradigm broadly encompasses Markov Chain Monte Carlo (MCMC) sampling, latent space search via Variational Autoencoders (VAEs), and evolutionary algorithms. Among the latter, Mol~GA~\cite{mol_GA} serves as a formidable baseline on the PMO benchmark~\cite{pmo} by evolving a population of molecular graphs. However, it incurs a high inference-time cost, typically requiring hundreds to thousands of oracle evaluations per instance. Similarly, the discrete diffusion model GenMol~\cite{genmol} effectively operates as an instance optimizer during goal-directed tasks. Although backed by a generalist denoiser pretrained on a billion-scale molecule dataset, for property optimization it relies on an iterative ``fragment remasking'' loop, a process functionally equivalent to Gibbs sampling. This enforces a dependence on active inference-time search, thereby preventing single-pass optimization.

\subsection{Amortized Optimization}
We explicitly distinguish amortized \textit{optimization} from amortized \textit{sampling}. While models such as Variational Autoencoders (VAEs) and discrete diffusion models learn amortized sampling distributions to generate novel molecules within the pretraining manifold, they function primarily as distribution learners. To perform property-optimization, which may require extrapolating to high-reward regions outside the training distribution, these models are typically coupled with active inference-time search mechanisms (e.g., latent space traversal for VAEs, or iterative fragment remasking loops in diffusion models like GenMol). This reliance effectively reverts them to instance optimizers.

In contrast, amortized optimization methods learn a conditional policy $\pi_\theta(y|x)$ to generate optimized structures directly, eliminating the need for extensive search or oracle querying at inference time. To ensure chemical validity within this paradigm, frameworks like LibINVENT~\cite{libinvent} (within REINVENT 4)~\cite{reinvent4} employ penalty-based regularization anchored to a pretrained prior. While effective for validity, this regularization can actively resist the policy drift required to reach high-reward regions far from the training distribution. Conditional encoder-decoder models, such as DrugEx v3~\cite{drugex_v3}, address this by training a Graph Transformer via multi-objective REINFORCE. To manage the high variance of policy gradients, DrugEx v3 relies on a fixed exploration network and Pareto-based ranking for multi-objective optimization.

The self-improvement approach GraphXForm~\cite{graphxform} trains a policy to imitate trajectories discovered by a search heuristic (TASAR~\cite{tasar}), which itself requires frequent oracle calls during training. The resulting policy is amortized at inference time, but the training loop remains oracle-intensive. On the other hand, AMORTIX optimizes the policy directly using reinforcement learning, stabilized by group-relative reward centering.

\section{Method}\label{sec:methodology}

\subsection{Structurally-Constrained Generation}
Structurally-constrained molecular design encompasses a range of tasks (scaffold decoration, motif extension, linker design, among others; see
Appendix~\ref{app:constrained_tasks} for definitions). We treat these uniformly: the
generation process initiates from a non-empty input subgraph $s_0$ (e.g., a molecular
scaffold or a fragment), and the policy $\pi_\theta$ sequentially adds atoms
and bonds to elaborate this structure. By setting $s_0$ to a single atom,
the formulation also encompasses \textit{de novo} generation. The goal is to
maximize a scalar reward $R(G)$ from an external oracle that evaluates the
desirability of the completed molecular graph $G$.

\subsection{Policy Architecture}

To implement the conditional policy $\pi_\theta(y|x)$, we utilize a decoder-only Graph Transformer that operates directly on the molecular graph representation and employs a step-wise action space~\cite{graphxform}. For training, we propose a direct on-policy RL approach built upon GRPO~\cite{grpo}. We provide a schematic overview of the hierarchical action space and training mechanism in Figure~\ref{fig:grxfrom}.

\paragraph{Action Space.}
To handle the discrete nature of molecule generation, the model constructs the graph via a sequential Markov Decision Process. At each step $t$, the policy samples a composite action decomposed into three
hierarchical levels: Level~0 \textit{Operation Selection} (Add Atom, Modify
Existing, or Stop), Level~1 \textit{Target Selection} (where to connect), and
Level~2 \textit{Bond Specification} (bond order). A validity mask $\mathcal{M}(s_t)$ is applied at each step to strictly enforce valence constraints, masking any actions that would violate basic atomic valency rules, thus ensuring that all generated structures are chemically valid by construction.

\paragraph{Initialization.}
Prior to RL fine-tuning, the model is pretrained on the ChEMBL database (version 35)~\cite{chembl} via supervised teacher forcing. We distinguish here between \textit{chemical validity} and \textit{plausibility}: while our action masking strictly enforces valence rules (validity), not all topologically valid graphs correspond to stable or realistic molecules. 
Pretraining imparts a prior of chemical plausibility, biasing the policy
toward stable, realistic molecules rather than arbitrary valid graphs.

We refer the reader to Figure~\ref{fig:architecture} and Appendix~\ref{model_architecture} 
for a detailed breakdown of the action space decomposition, the model architecture 
including the attention mechanisms, and the pretraining.

\subsection{Group Relative Policy Optimization (GRPO)}
\label{sec:grpo}
A key contribution is the adaptation of GRPO to the domain of constrained molecular design. Standard RL methods often struggle in this domain due to the high variance in conditional difficulty.
\paragraph{Motivation.}
As established in Section~\ref{sec:intro}, the heterogeneous difficulty of starting
structures produces reward signals on incomparable scales (see Figure~\ref{fig:grpo} in Appendix \ref{app:finetuning} for an
illustration). GRPO addresses this by centering rewards within groups of trajectories sampled
from the same starting structure, which removes the need for a learned value critic.

\paragraph{Group Formation and Update.}
For each training step, we sample a batch of $B$ starting structures and
generate $G$ completions per structure via independent sampling from the
policy. We adopt the Dr.\ GRPO formulation~\cite{dr_grpo}, computing the
advantage $A_{i,j}$ by centering each reward against the group mean:
\begin{equation}
    A_{i,j} = R(O_{i,j}) - \mu_i, \quad
    \mu_i = \frac{1}{G}\sum_{k=1}^G R(O_{i,k})
\end{equation}
This formulation makes two deliberate departures from the original GRPO objective. First, the division by group standard deviation is omitted: structures where all
completions score similarly (either trivially easy or near-impossible) would
otherwise receive disproportionate weight in the gradient. Second, the normalization by trajectory length is omitted, since it biases the update by completion length: shorter completions receive larger per-step weight
when the advantage is positive, and longer ones are penalized less when it is
negative. Since sampling
trajectories is inexpensive relative to updating the policy parameters, we
perform a single gradient step per batch, making PPO-style clipping
unnecessary. The gradient estimator aggregates contributions across all
generation steps $t$:
\begin{equation}
\nabla_\theta \mathcal{J}(\theta) \approx \frac{1}{BG} \sum_{i=1}^B
\sum_{j=1}^G A_{i,j} \sum_{t=1}^{T_{i,j}} \nabla_\theta \log
\pi_\theta(a_t | s_{<t})
\end{equation}
where $a_t$ is the action at step $t$ given partial graph $s_{<t}$. The training procedure is summarized in Appendix \ref{app:finetuning} (Algorithm~\ref{alg:AMORTIX_inner}).

\section{Experiments}

Our experiments focus on structurally-constrained optimization, where a policy
is trained once and then applied to many unseen starting structures. We first
evaluate goal-directed scaffold decoration on held-out scaffolds, comparing
against amortized and instance-optimization baselines and ablating the training
algorithm across 23 objectives (Section~\ref{subsec:ablation}). We then examine how
group-relative reward centering affects training stability and sample diversity.
Finally, we demonstrate few-shot rule transfer via a case study on prodrug design (Section~\ref{sec:prodrug}).
Results on the PMO benchmark, which consists of \textit{de novo} molecular design tasks without any input starting structures, are reported in Appendix~\ref{pmo}.

\subsection{Goal-Directed Scaffold Decoration}\label{subsec:kinase}

Scaffold decoration attaches functional groups and side chains to a fixed core molecular structure (the scaffold). This evaluation tests whether an amortized policy can elaborate unseen scaffolds into valid biological inhibitors.

\paragraph{Setup.}
We use ZINC-250k~\cite{zinc} as the scaffold source. To enforce generalization, we perform Murcko scaffold decomposition
\cite{murcko, rdkit} and cluster scaffolds via the Butina
algorithm~\cite{butina} on Morgan fingerprints
(\cite{morgan, tanimoto}, Tanimoto cutoff $0.4$). Entire clusters
are assigned to train, validation, or test sets, reserving 500
scaffolds for testing; we repeat this procedure for three random splits, train each model once per
split, and report mean $\pm$ SD over the three splits. Test scaffolds are thus topologically distinct from training
(Figure~\ref{fig:tsne}). Validation scaffolds are used only to log training progress and not for early stopping; the oracle calls this requires are therefore not counted towards the training budget. Oracle calls are counted per distinct molecule: within one training run for amortized methods, and within the search for one starting scaffold for instance optimizers.

\paragraph{Objectives.}
We evaluate three independently trained tasks: single-objective
optimization for GSK3$\beta$ and JNK3, and a Kinase Multi-Parameter
Optimization (MPO) task. Following~\cite{rationale_RL}, the MPO
reward aggregates four components: GSK3$\beta$ and JNK3 activity
(Random Forest classifiers from the Therapeutics Data Commons,
TDC~\cite{tdc}), QED (Quantitative Estimate of Drug-likeness)~\cite{qed}, and SA (Synthetic Accessibility)~\cite{sa_score} normalized as
$\text{SA}' = (10 - \text{SA})/9$; the latter two metrics are included to ensure the generated candidates remain drug-like and synthetically tractable:
\begin{equation}\label{eq:kinase}
    R(M) = \tfrac{1}{4}\!\left(P_{\text{GSK3}\beta} + P_{\text{JNK3}}
    + P_{\text{QED}} + P_{\text{SA}'}\right).
\end{equation}

\paragraph{Baselines and Ablations.}
We compare against amortized baselines that natively support
structural elaboration: LibINVENT~\cite{libinvent,reinvent4},
GraphXForm~\cite{graphxform}, and DrugEx~v3~\cite{drugex_v3}; and
against instance optimizers Mol~GA~\cite{mol_GA} and
GenMol~\cite{genmol}. We exclude primarily \textit{de novo}
optimizers that do not natively support hard substructure
constraints, ensuring a fair comparison on the constrained-generation
task. All amortized methods start from a ChEMBL-pretrained prior: AMORTIX and
GraphXForm use the same filtered ChEMBL~35 set, while LibINVENT and DrugEx~v3
start from their publicly released ChEMBL priors
(Appendix~\ref{app:pretraining}). ZINC-250k serves only as the source of
scaffolds. Amortized models train within $\approx$50{,}000 oracle calls (AMORTIX,
LibINVENT) to $\approx$640{,}000 (DrugEx v3) and generate one completion per test scaffold.
Instance optimizers are run with a budget of up to $10{,}000$ calls
\textit{per scaffold} for Mol~GA (20\% of AMORTIX's entire training
budget for a single test case) and $\approx$1{,}000 calls for GenMol.
When comparing cumulative cost, we charge instance optimizers only for the
calls needed to \textit{converge}, i.e., until their mean best-so-far score is
within $0.01$ of its final value (Appendix~\ref{app:oracle_analysis}). We additionally evaluate three
AMORTIX ablations: \textbf{AMORTIX-DeNovo} (trained only from a single carbon atom, and given the
test scaffold as its initial graph at inference), \textbf{AMORTIX-REINFORCE} (global
baseline), and \textbf{AMORTIX-PPO} (learned value critic). 
Implementation details are in Appendix~\ref{app:hyperparams}.

\begin{table}[b!]
    \caption{Optimization scores across three tasks (mean $\pm$ SD over 3 scaffold
    splits). \textbf{AMORTIX} denotes the default GRPO variant. Amortized methods generate one molecule per test scaffold; instance
    optimizers report the best molecule found per scaffold.}
    \label{tab:main_performance}
    \begin{center}
        \begin{small}
        \begin{sc}
            \begin{tabular}{lccc}
                \toprule
                Method & GSK3$\beta$ & JNK3 & Kinase MPO \\
                \midrule
                \multicolumn{4}{l}{\textit{Amortized Policies}} \\
                LibINVENT & $0.414 \pm 0.122$ & $0.261 \pm 0.065$ & $0.453 \pm 0.007$ \\
                GraphXForm & $0.110 \pm 0.001$ & $0.051 \pm 0.002$ & $0.409 \pm 0.003$ \\
                DrugEx v3  & $0.194 \pm 0.004$ & $0.356 \pm 0.003$ & $0.354 \pm 0.003$ \\[1ex]
                AMORTIX-DeNovo & $0.572 \pm 0.015$ & $0.381 \pm 0.029$ & $0.397 \pm 0.005$ \\
                AMORTIX-REINFORCE & $0.757 \pm 0.084$ & $0.754 \pm 0.092$ & $0.475 \pm 0.120$ \\
                AMORTIX-PPO & $0.725 \pm 0.017$ & $0.728 \pm 0.008$ & $0.552 \pm 0.015$ \\
                \textbf{AMORTIX} & \textbf{0.775 $\pm$ 0.005} & \textbf{0.796 $\pm$ 0.004} & \textbf{0.619 $\pm$ 0.004} \\
                \midrule
                \multicolumn{4}{l}{\textit{Instance Optimizers}} \\
                Mol~GA & $0.735 \pm 0.014$ & $0.659 \pm 0.011$ & $0.461 \pm 0.013$ \\
                GenMol & $0.700 \pm 0.012$ & $0.470 \pm 0.015$ & $0.479 \pm 0.012$ \\
                \bottomrule
            \end{tabular}
        \end{sc}
        \end{small}
    \end{center}
\end{table}

\paragraph{Results.}
Table~\ref{tab:main_performance} reports objective scores across
the three tasks. AMORTIX achieves the highest score on all three,
outperforming both amortized and instance-optimization baselines while requiring no oracle calls at test time. Among amortized baselines, GraphXForm and DrugEx~v3 score low on the biological
targets; LibINVENT is stronger on most tasks but still trails AMORTIX by a large
margin on both GSK3$\beta$/JNK3 ($0.41$/$0.26$ vs.\ $0.78$/$0.80$) and Kinase MPO
($0.45$ vs.\ $0.62$).

The comparison against instance optimizers reveals a clearer
pattern. On
the single-objective tasks Mol~GA and GenMol trail AMORTIX moderately on GSK3$\beta$ and
clearly on JNK3, but on Kinase MPO they fall well behind.
The sub-task breakdown
(Table~\ref{tab:mpo_breakdown_appendix}, Appendix~\ref{app:mpo_breakdown}) shows that on the Kinase MPO task, both methods exploit the easy heuristic components (Mol~GA QED~$0.81$;
GenMol QED~$0.88$, SA~$2.93$) and fail to penetrate the sparse
biological-target manifolds, scoring below $0.24$ on
GSK3$\beta$/JNK3. AMORTIX makes the opposite tradeoff: it prioritizes the difficult biological
targets ($0.74$/$0.59$) at a substantial cost in QED ($0.38$ vs.\ $0.81$--$0.88$).
Setting a QED floor in the reward could address this.

The \textit{de novo} variant is trained like AMORTIX but always starts from a single
carbon atom; at test time it receives the scaffold as its initial graph, like
all other variants. It clearly underperforms on all three tasks, indicating that
training on diverse scaffold starting structures is helpful in teaching the policy to
optimize well. We analyze the two training-algorithm ablations,
REINFORCE with a global baseline and PPO with a value critic, across 23
objectives in Section~\ref{subsec:ablation}.

\begin{figure}[h]
    \centering
    \includegraphics[width=0.84\linewidth]{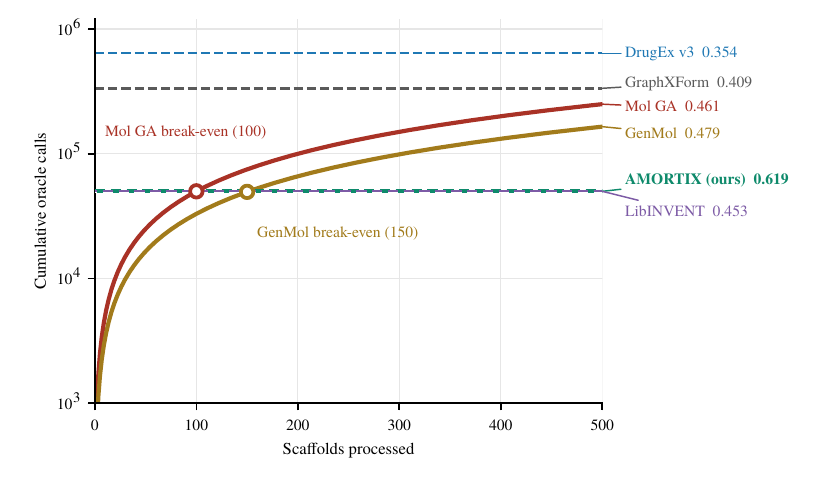}
    \caption{\textbf{Cost vs.\ performance on Kinase MPO.} Cumulative oracle calls per method as test
scaffolds are processed; numbers next to each curve are mean test-set objective scores.
Amortized methods front-load cost into a one-time training budget and use no oracle calls at inference. Instance optimizers pay per scaffold; we charge them the number of oracle calls
until their mean best-so-far score over all test scaffolds converges (within
$0.01$ of its final value): $\approx$500 for Mol~GA and $\approx$330 for GenMol
(Table~\ref{tab:oracle_calls}).
Open circles mark where Mol~GA ($\approx$100 scaffolds) and GenMol ($\approx$150 scaffolds)
match AMORTIX's training budget.}
    \label{fig:oracle_cost}
\end{figure}

Beyond raw performance, AMORTIX's one-time training budget of
$\approx$50{,}000 oracle calls pays for itself quickly. On Kinase MPO, Mol~GA
needs $\approx$500 calls per scaffold to converge and GenMol $\approx$330, so
their cumulative cost overtakes AMORTIX's after roughly $100$ and $150$ test
scaffolds, respectively (Figure~\ref{fig:oracle_cost}). On GSK3$\beta$ and JNK3
the break-even comes earlier still, because both need more calls per scaffold
($\approx$5{,}000--7{,}000 for Mol~GA, $\approx$800--900 for GenMol), which puts
it between about $7$ and $65$ scaffolds (Figures~\ref{fig:molga_convergence}
and~\ref{fig:genmol_convergence} in Appendix~\ref{app:oracle_analysis}). At 500
scaffolds, the instance optimizers have consumed $\approx$250{,}000 and
$\approx$165{,}000 calls on Kinase MPO, roughly five and three times AMORTIX's allowed budget, respectively.

\subsection{Ablation of the Training Algorithm}\label{subsec:ablation}
To test the training algorithm more broadly, we reuse the 22 PMO scoring
functions as scaffold-decoration rewards on the same held-out scaffolds, i.e., 23 total objectives including Kinase MPO. We train AMORTIX (using GRPO, REINFORCE, and PPO variants) and additionally LibINVENT under the same budget of $\approx$50{,}000 oracle calls
(Table~\ref{tab:ablation_summary}; per-objective scores are in
Appendix~\ref{app:ablation_full}).

\begin{table}[h]
\caption{Summary of the 23-objective ablation. Mean score is averaged over
objectives. Split SD is the standard deviation across 3 scaffold splits, averaged over
objectives. Further details are in Appendix~\ref{app:ablation_full}.}
\label{tab:ablation_summary}
\begin{center}
\begin{small}
\begin{sc}
\begin{tabular}{lcccc}
\toprule
Method & Mean score & Best on & Split SD & GRPO higher on \\
\midrule
GRPO & 0.570 & 18/23 & 0.004 & --- \\
PPO & 0.527 & 5/23 & 0.013 & 18/23 \\
REINFORCE & 0.512 & 0/23 & 0.072 & 22/23 \\
LibINVENT & 0.320 & 0/23 & 0.021 & 23/23 \\
\bottomrule
\end{tabular}
\end{sc}
\end{small}
\end{center}
\end{table}

AMORTIX with GRPO achieves the best score on most of the objectives, and LibINVENT scores below it on all of them. On most individual objectives, GRPO's margin over the REINFORCE and PPO baselines is small, and the large gains are
concentrated on a few objectives, notably DRD2 and Perindopril MPO, where both baselines fall far behind. The more notable difference is consistency: averaged over the objectives, GRPO has $17\times$ and $3\times$ lower standard deviation across scaffold splits than REINFORCE and PPO respectively. This consistency is particularly important because our goal is a policy that is trained once and then applied to unseen molecular structures without further oracle access, so a poor run cannot be compensated for at inference time.

PPO is the stronger of the two baselines. Its learned critic gives more
consistent results than a global baseline and recovers some of the gap in score,
and on five objectives it matches or slightly beats GRPO. Where it falls
behind, it does so by a wide margin, for example on DRD2, Perindopril MPO and
Mestranol Similarity.

We note that GRPO does not gain its consistency by significantly narrowing the
output distribution. Averaged over the 23 objectives, the diversity of the
generated molecules is close to that of both baselines
(Table~\ref{tab:diversity}, Appendix~\ref{app:diversity}), although GRPO is somewhat less diverse than
REINFORCE within a scaffold, notably in uniqueness on GSK3$\beta$ and JNK3. The
clearest difference is on Kinase MPO, where REINFORCE collapses onto a handful
of repeated molecules while GRPO continues to produce many distinct ones.

\begin{figure}[h]
    \centering
    \includegraphics[width=0.817\linewidth]{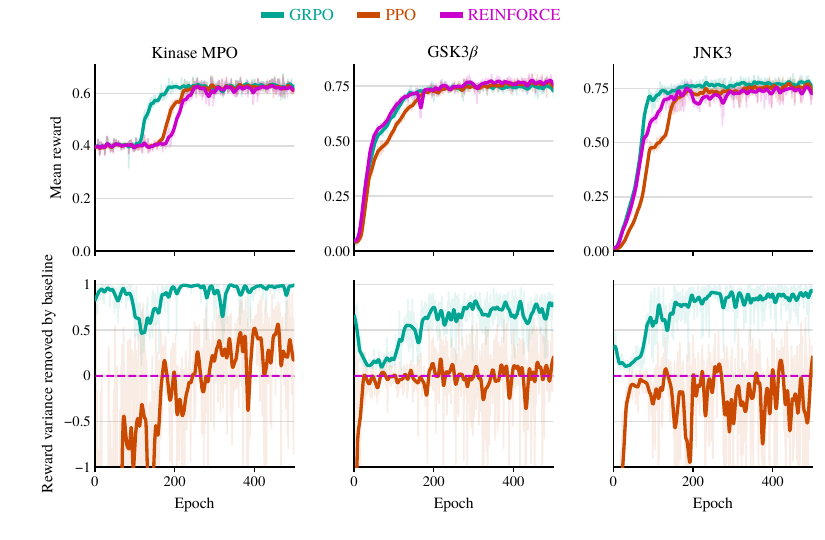}
    \caption{\textbf{Training signal on three scaffold-decoration tasks.}
\emph{Top:} mean training reward. \emph{Bottom:} fraction of the reward variance
removed by the baseline (defined in the text); the dashed line is the global
baseline. Thin lines: per epoch; thick lines: Gaussian-smoothed ($\sigma=3$
epochs); values below $-1$ are clipped. One run per method on a single scaffold
split.}
    \label{fig:variance}
\end{figure}

Figure~\ref{fig:variance} shows the training signal of the three variants. All of them reach about the same mean reward on sampled completions of the training scaffolds, with GRPO getting
there earlier on Kinase MPO and JNK3; the difference lies in how much of the
reward variation reaches the update. To quantify this, we measure for each
training batch the fraction of the reward variance that the baseline removes from
the advantage, $1-\mathrm{Var}\text{(Advantage)}/\mathrm{Var}\text{(Reward)}$, over the trajectories started
from scaffolds and before any normalization (bottom row). For a global baseline
this is zero by construction, since subtracting a constant does not change the
variance. For GRPO it is the share of the reward variance
that comes from differences between scaffolds. For PPO the baseline is the
critic's value of the starting structure, and a negative value means that the
critic adds variance. GRPO removes most of the variance once
the policy has improved: averaged over the second half of training, about 95\% on
Kinase MPO, 86\% on JNK3 and 71\% on GSK3$\beta$, so that differences between
scaffolds account for most of the variation of the rewards. On the two single-objective tasks this share dips to about 10\% early in
training, while the reward improves fastest, and grows afterwards; on Kinase MPO
it dips to about 50\%. The dip comes from the variation among samples of the
same scaffold, which is largest while the policy is learning, whereas the
variance between scaffolds changes little around that point
(Appendix~\ref{app:ablation_full}). The critic of PPO removes little:
averaged over the second half of training the fraction of reward variance it removes is about 23\% on Kinase
MPO and about zero on GSK3$\beta$, and on JNK3 it is negative ($-21\%$). During
the first epochs it adds variance on all three tasks. The unnormalized advantage
spread of all three variants is shown in Figure~\ref{fig:adv_sd_raw} (Appendix~\ref{app:ablation_full}).

\subsection{Case Study: Prodrug Transfer for Blood--Brain-Barrier Penetration}
\label{sec:prodrug}

As a case study in practical applicability, we consider prodrug design to improve
blood--brain-barrier (BBB) penetration, which limits the access of many drugs to
the central nervous system. A prodrug~\cite{prodrug1} is a chemically modified form of a
drug (the parent) that is converted back to the parent in the body. This lets a drug gain properties it lacks on its own, such as the ability to reach the brain, while the parent remains the active compound~\cite{prodrug2}. We test
whether AMORTIX, fine-tuned on just four exemplar drugs, can apply learned
modification rules to 52 unseen parents at inference time, without further
training. The reward (Eq.~\ref{eq:prodrug_reward}) combines a predicted property
term, a drug-likeness gate, and a structural cleavability term:
\begin{equation}
  R(G \mid S_0) = P_{\text{BBB}}(G) \cdot
  \mathbf{1}\!\left[\mathrm{QED}(G) \ge 0.4\right] \cdot
  \rho(G, S_0),
  \label{eq:prodrug_reward}
\end{equation}
where $P_{\text{BBB}}$ is a predicted BBB penetration
probability from the MiniMol foundation model~\cite{minimol} fine-tuned
on the TDC \texttt{BBB\_Martins} benchmark dataset~\cite{martins, moleculenet};
at the time of writing this is the state-of-the-art model on this
task~\cite{leaderboard}. The term $\rho(G, S_0) \in [0,1]$ is
the \emph{junction purity}: the fraction of new bonds between parent
and addon that belong to a cleavable ester or carbamate class. The
multiplicative form ensures that a high BBB score cannot compensate for a non-cleavable or drug-unlike generation.

\paragraph{Setup and results.}
We fine-tune the policy on four parent drugs (morphine, GABA, nipecotic acid, aspirin) and, additionally, a single carbon atom as starting points, initializing with the ChEMBL-pretrained checkpoint and training using GRPO with 32 samples per parent for 50 epochs (a maximum of $8{,}000$ unique reward evaluations in total). We evaluate on 52 held-out drugs spanning multiple therapeutic classes (listed in
Appendix~\ref{app:prodrug}), sampling 32 candidates per parent ($1{,}664$ in total). The best candidate per parent is chosen by scoring all 32 with
the reward, which is inexpensive to compute here. Representative cases are shown in
Figure~\ref{fig:prodrug_main}, with further examples in
Appendix~\ref{app:prodrug}.

\begin{figure}[H]
  \centering
  \includegraphics[width=\linewidth]{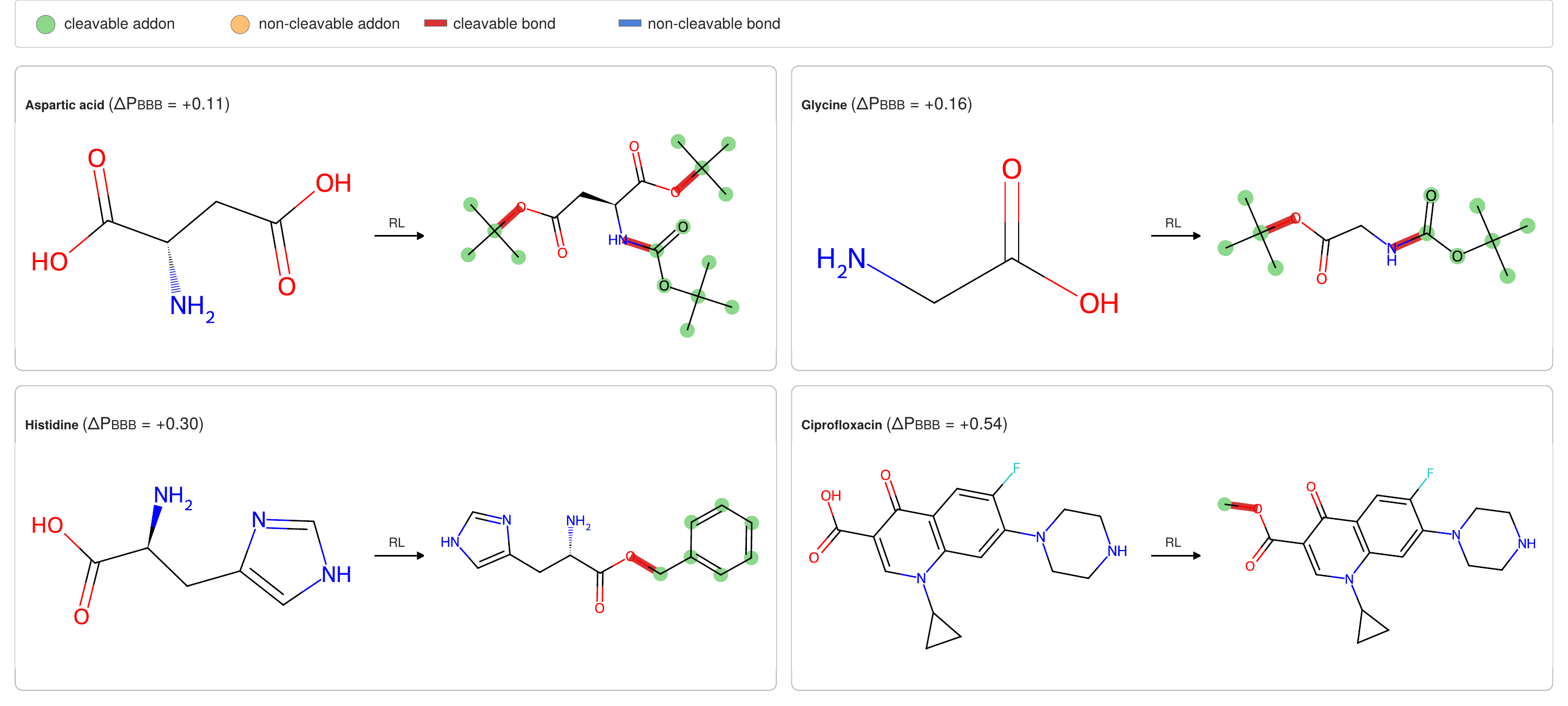}
  \caption{\textbf{Pure prodrugs generated for unseen parent drugs.} Four of the
  16 pure prodrugs (every parent--addon bond cleavable, parent skeleton
  unmodified), each shown next to its parent, covering three cleavable groups
  found (alkyl ester, alkyl carbamate, benzyl ester). Green: addon atoms
  released on hydrolysis; red: cleavable bond. $\Delta P_{\text{BBB}}$:
  predicted BBB-penetration lift over the parent. The other 12 are in
  Appendix~\ref{app:prodrug}.}
  \label{fig:prodrug_main}
\end{figure}

The policy recovers three distinct cleavable motifs (alkyl ester, alkyl carbamate, and benzyl ester). Of the 52 best candidates, 16 (31\%) are pure prodrugs
(every parent--addon junction cleavable, parent skeleton unmodified) and 33
(63\%) carry at least one cleavable handle, with a mean predicted
BBB-penetration lift of $+0.26$ over the parent. In all 16 pure prodrugs the parent's carboxylic acid is esterified; for Aspartic
acid, Glycine and Cysteine the amine is also converted to a carbamate, whereas
the thiol of Cysteine and the phenols of Tyrosine and 5-HTP are left unmodified. The reward distribution is given in Appendix~\ref{app:prodrug}. The
MiniMol BBB score is a predictive proxy, and we make no claim about in vivo BBB penetration or brain exposure. This is a single-method case study, and comparisons with other approaches remain future work.

\section{Limitations}
While AMORTIX performs strongly on structurally-constrained tasks, several
limitations remain. It works with 2D molecular graphs and does not explicitly
model 3D conformations, which matters for binding
affinity in practice. Its action space is restrictive as well: generation is
purely additive, and the action masking assumes a single, connected initial
molecule. Scaffold hopping, more flexible core modifications and fragment
linking would therefore require atom deletions or substitutions, along with
changes to the constraint logic. Pretraining also fixes a single canonical construction order for each molecule, and whether using randomized orders helps was not tested.

Our evaluation is also limited in scope. All oracles are cheap proxies
(random-forest activity classifiers, a QED/SA mixture, and a predicted
blood-brain-barrier probability), and we have not tested AMORTIX with docking or
other 3D-aware oracles, where each call is far more expensive and the spread of
achievable scores across scaffolds may differ.
Since instance optimizers need a separate search for every starting structure,
comparing against them at scale is costly, so we evaluate them on a small number
of objectives. Likewise, the prodrug study is a single-method case study, and
comparisons with baselines such as LibINVENT fine-tuned with specified exit
vectors, or GenMol, remain to be done.

Finally, although inference is cheap, training is still oracle-intensive and
typically requires thousands of calls to converge to a robust policy. Our
break-even analysis counts only oracle calls, which is the relevant cost when
oracles are expensive. When oracles are cheap, as for the proxies used here,
wall-clock time matters as well: one rollout with AMORTIX took on average
0.05--0.27 seconds per input scaffold, and a training run took about 4--5
hours on one GPU, roughly the time GenMol needed to optimize 10--15 scaffolds
(Appendix~\ref{app:oracle_analysis}). Whether and to what extent amortization
pays off in practice remains an open question for commonly used expensive oracles
and real-world molecule design scenarios.

\section{Conclusion}
AMORTIX demonstrates that the computational burden of molecular optimization can be effectively shifted from inference-time search to training-time learning. Our key conceptual insight is that when training an RL policy for molecular generation, we must account for the high variance in the difficulty of successful molecular generation across different initial structures. We address this by adapting GRPO; our ablations show that it yields higher and markedly more consistent scores than REINFORCE with a global baseline or PPO with a value critic. By enforcing chemical validity through the model architecture and stabilizing the learning process with group-relative baselines, we achieve a policy capable of fast, generalized optimization. This approach offers a scalable alternative to iterative search methods, enabling high-throughput design without the prohibitive cost of repeated oracle evaluations. Looking forward, we plan to evaluate the framework on tasks requiring atom and bond removal (e.g., scaffold morphing, lead optimization) and linking disjoint fragments. Finally, we see substantial potential in hybrid approaches that combine the strengths of instance optimization and amortization. By using a robust amortized policy to rapidly map high-value regions of chemical space across various substructural and synthesis constraints, iterative search algorithms could be initialized with near-optimal candidates, enabling the rigorous local refinement of instance optimization at a fraction of its traditional computational cost.

\begin{ack}
Muhammad bin Javaid, Berke Ki\c{s}in, Alexander Mitsos, and Martin Grohe were supported by the Werner Siemens Foundation within the WSS project of the century ``catalaix''. Ashima Khanna-Reiter, Jonathan Pirnay, and Dominik G. Grimm were supported by the Deutsche Forschungsgemeinschaft (DFG, German Research Foundation) grant 466387255, Berke Ki\c{s}in and Martin Grohe by grant GR 1492/20-2, and Alexander Mitsos by grant 466417970 within the Priority Programme ``SPP 2331: Machine Learning in Chemical Engineering''. Computations were performed with computing resources granted by RWTH Aachen University under project rwth1945.
\end{ack}

\bibliography{references} 
\bibliographystyle{unsrt}
\newpage
\appendix
\onecolumn 

\section{Impact Statement}\label{app:impact}
This work advances the field of automated drug discovery by introducing a computationally efficient amortized optimization framework. By removing oracle queries and iterative search at inference time, our approach can reduce the computational cost (and thus the energy footprint) of molecular design at scale, although training itself remains compute-intensive. Given the potential for dual-use in generative chemistry, we emphasize the necessity of human oversight and ethical due diligence to prevent the design of harmful compounds.

\section{Taxonomy of Generative Optimization}\label{appendixA}

\begin{figure}[h]
\centering
\includegraphics[width=\linewidth]{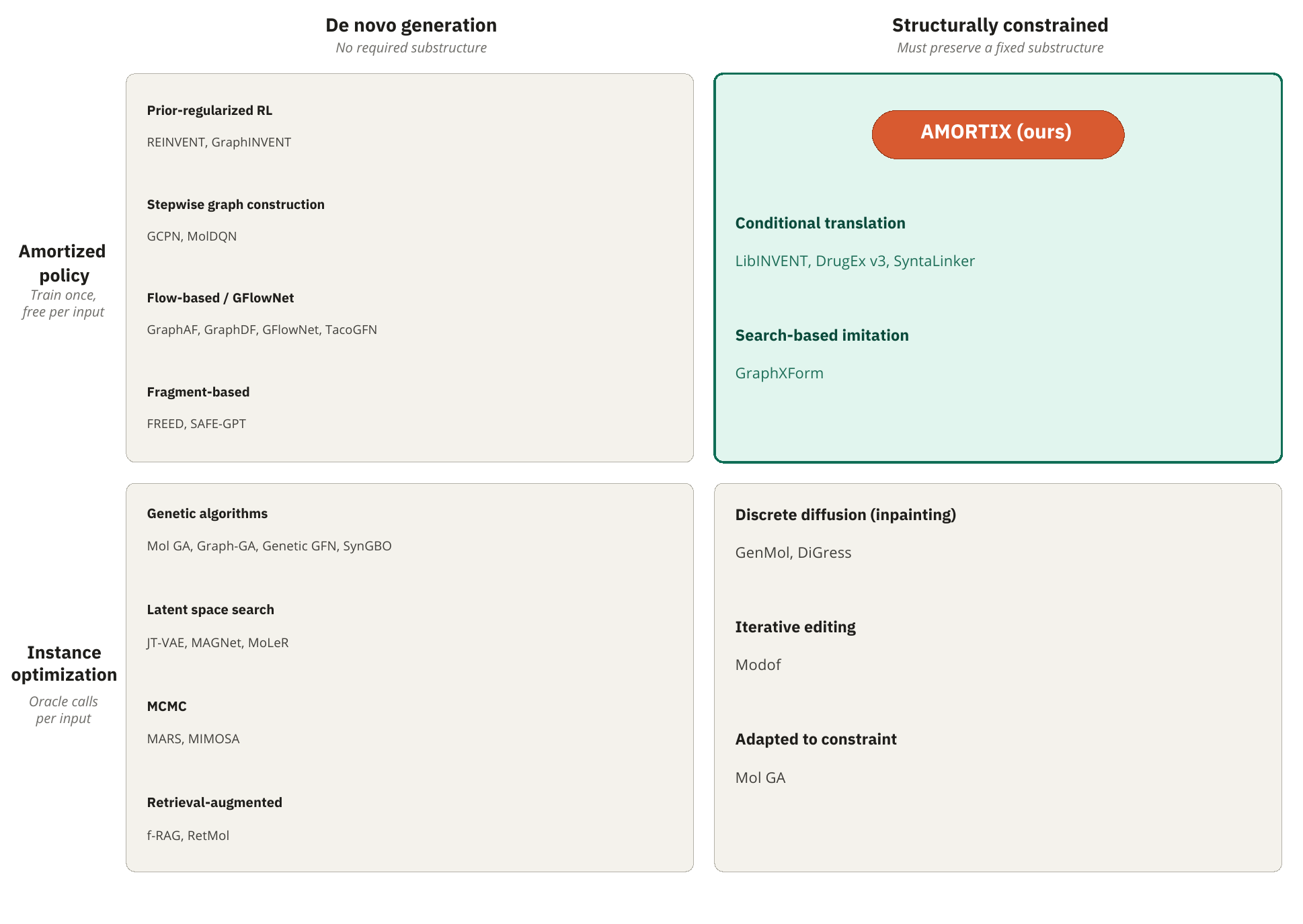}
\caption{Taxonomy of generative molecular optimization methods, organized by amortization (rows) and structural constraint (columns). The amortized + structurally-constrained quadrant (highlighted) is comparatively underserved: existing methods in this regime either rely on expensive search-based imitation (GraphXForm), struggle to scale across diverse scaffolds (LibINVENT), or use global rewards that do not handle heterogeneous scaffold difficulty (DrugEx v3). AMORTIX targets this regime. Mol~GA can be applied to constrained tasks via a hard scaffold-preserving penalty; it is not natively scaffold-aware.}
\label{fig:taxonomy}
\end{figure}

We structure our analysis of the generative molecular design landscape 
through the lens of two distinct optimization paradigms: \emph{instance 
optimization}, which treats every design task as a unique search problem 
to be solved from scratch, and \emph{amortized optimization}, which 
distills optimization logic into a learned policy which can be re-used. 
A second axis cuts across both paradigms: whether the method natively 
supports \emph{structurally-constrained} generation (preserving a fixed 
input substructure) or only \emph{de novo} design. Figure~\ref{fig:taxonomy} 
places existing methods along these two axes; the remainder of this 
section elaborates on each category.

\subsection{Instance Optimization}
Methods in this category treat molecular design as a combinatorial optimization problem. They rely on an external oracle (scoring function) to guide a traversal of chemical space during the inference phase. While often effective at finding high-scoring molecules, they suffer from high computational cost and a lack of transferable knowledge.

\textbf{Genetic Algorithms and MCMC.} Evolutionary strategies remain formidable baselines in drug discovery. Methods such as Graph-GA and Mol~GA~\cite{graphGA, mol_GA} evolve a population of molecular graphs through crossover and mutation operations. These classical frameworks often outperform deep learning methods, with Mol~GA achieving good performance in sample efficiency tasks such as the PMO benchmark. However, they are computationally expensive, typically requiring thousands of oracle evaluations per target task to effectively navigate the chemical space, making them impractical for high-throughput scenarios or expensive biological oracles. Similarly, Markov Chain Monte Carlo (MCMC) approaches like MARS~\cite{mars} and MIMOSA~\cite{mimosa} navigate chemical space by iteratively editing molecules (adding/deleting bonds or atoms) via a proposal network. As these methods optimize a single instance via repeated sampling steps, they are operationally distinct from amortized policies.

\textbf{Latent Space Search (VAEs).}
Deep generative models based on Variational Autoencoders (VAEs), such as JT-VAE~\cite{jtvae}, MoLeR~\cite{moler}, MAGNet~\cite{magnet} and PS-VAE~\cite{psvae}, map discrete molecular graphs to a continuous latent space. To perform optimization, these frameworks rely on inference-time search algorithms such as Bayesian Optimization, Molecular Swarm Optimization~\cite{mso} or gradient ascent on a learned property predictor to traverse the latent manifold and decode new candidates. Consequently, the ``generative'' step is a decoding operation for an underlying search process. Similarly, deep editing models like Modof~\cite{modof} are trained to predict single-step structural modifications (e.g., disconnection and replacement). To achieve significant structural elaboration, such models must be applied iteratively in a loop, effectively reverting to a search-based instance optimization paradigm rather than single-pass generation.

\textbf{Discrete Diffusion and Iterative Refinement.}
Recently, discrete diffusion models have been applied to molecular graphs~\cite{digress} and sequences~\cite{diffusion_sequences}. While such frameworks technically support molecule completion (inpainting) by fixing scaffold nodes during the reverse process, we exclude them as direct baselines because they function primarily as distribution learners rather than optimizers. Conditional sampling $p_\theta(x|\text{scaffold})$ is strictly bound by the quality of the training data; in the absence of dense, high-reward examples in the training set (the exploration gap), standard diffusion lacks the extrapolation capability to discover novel, high-performing solutions.

To bridge this gap, state-of-the-art approaches like GenMol~\cite{genmol} must reintroduce inference-time search. GenMol serves as a generalist model using Sequential Attachment-based Fragment Embedding (SAFE)~\cite{safe} with a non-autoregressive, bidirectional parallel decoding scheme. However, to perform goal-directed tasks like lead optimization, it relies explicitly on an iterative ``fragment remasking'' loop. As the authors note, this process functions as a mutation operation where specific fragments are masked and regenerated iteratively based on oracle feedback. This effectively constitutes a guided random walk (Gibbs sampling) in the neighborhood of a seed molecule. Consequently, GenMol functions as an instance optimizer: it does not learn a direct, amortized mapping from a starting structure to an optimized molecule, but rather relies on a computationally expensive denoising kernel to actively search chemical space at test time.

\textbf{Retrieval-Augmented Generation.} Retrieval-augmented methods address the rigidity of fixed fragment vocabularies (see Vocabulary-Based Methods below) by introducing inference-time retrieval loops, a step that shifts them towards instance optimization. RetMol~\cite{retmol} employs a ``soft'' retrieval paradigm, fetching whole exemplar molecules to fuse into the generator's latent space. It relies on an iterative refinement procedure: generating candidates, evaluating them with the oracle, and updating the retrieval database dynamically. Similarly, f-RAG~\cite{frag} utilizes a dual-retrieval mechanism (fetching both ``hard'' structural anchors and ``soft'' latent guidance). By updating its fragment memory online during exploration, f-RAG effectively reintroduces a search component, using the learned policy to accelerate the discovery of high-scoring fragments that are then stored for subsequent retrieval.

\subsection{Amortized Optimization (Learned Policies)}

\textbf{Prior-Regularized RL.} To ensure chemical validity, frameworks like REINVENT~\cite{reinvent1,reinvent4} and GraphINVENT~\cite{graphINVENT} rely on anchored strategies (e.g., Direct Augmented Likelihood), which explicitly regularize the reward with a fixed pretrained prior. This prior serves to constrain the agent to the training distribution. While this helps ensure chemical validity, it introduces a trade-off: maximizing the reward often requires drifting far from the prior, which the regularization term actively resists. Consequently, achieving high scores on specific targets requires an iterative learning process to gradually shift the agent's distribution.

\textbf{Stepwise Graph Construction.} Early deep reinforcement learning approaches formulated molecular design as a Markov Decision Process (MDP) operating at the atom level. GCPN~\cite{gcpn} employs a Graph Convolutional Policy Network to construct molecules node-by-node and edge-by-edge, using Proximal Policy Optimization (PPO)~\cite{ppo} to update the policy. By treating valency checks as intermediate environment rewards, it learns to generate valid structures. MolDQN~\cite{moldqn} approaches the same MDP formulation through using Double Q-Learning to greedily select atoms and bonds that maximize the expected future reward.

\textbf{Flow-based Models.} Normalizing flows model the molecular distribution $p(x)$ by learning an invertible mapping between molecular graphs and a latent base distribution. To perform goal-directed optimization, these frameworks typically employ a two-stage paradigm: pretraining on a large dataset to learn chemical validity, followed by RL fine-tuning. GraphAF~\cite{graphaf} formulates generation as an autoregressive flow; however, to apply continuous dynamics to discrete graphs, it relies on dequantization (adding real-valued noise), which introduces stochasticity that hinders the precise optimization of property constraints. To resolve this, GraphDF~\cite{graphdf} replaces continuous transformations with discrete latent variables using invertible modulo shift transforms. By eliminating dequantization noise, GraphDF provides a more stable mapping for the RL agent, enabling more efficient navigation of the chemical property landscape.

\textbf{Generative Flow Networks (GFlowNets).} Generative Flow Networks (GFlowNets)~\cite{gflownet} are designed to learn a stochastic policy that samples candidates with probability proportional to their reward. While theoretically powerful for diverse mode coverage, prioritizing this diversity can trade off against the rapid peak-finding required in sparse, high-dimensional chemical spaces, as suggested by the relatively poor performance of GFlowNets on the PMO benchmark. To address this, hybrid methods such as Genetic GFN~\cite{genetic_gfn} augment training with evolutionary operators; however, this shifts the paradigm back toward instance optimization. By relying on a GA to populate the replay buffer with task-specific elite solutions, Genetic GFN focuses on solving the current instance rather than learning a generalizable policy for new constraints or scaffolds without restarting the online training loop. Crucially, we distinguish our approach from standard GFlowNets based on the structural requirements of the design task. The canonical GFlowNet formulation relies on a fixed root state $s_0$ (typically an empty graph). While architectures like TacoGFN~\cite{taco_gfn} solve target-conditional generation (e.g., conditioning on a protein pocket), they still initiate flow from the empty graph. To enforce a structural hard-constraint (e.g., the output must contain the subgraph $C$) in this setting, the agent is forced to reconstruct the subgraph $C$ from scratch. This introduces an exploration bottleneck, as the model must correctly navigate a long, specific trajectory simply to reproduce the input before optimization can begin. Alternatively, treating the scaffold itself as the starting root ($s_0 = C$) requires the framework to handle variable starting states; a non-trivial adaptation that effectively requires training a distinct flow for each unique scaffold (instance optimization) or fundamental architectural redesign. Consequently, we focus our comparison on paradigms that natively support the variable-starting point regime.

\textbf{Fragment-Based Methods.} To improve optimization efficiency and ensure local substructure validity, methods like FREED~\cite{freed} and SAFE-GPT~\cite{safe} elevate the action space from atoms to fragments. FREED sequentially attaches chemically valid fragments to a growing structure and uses Soft Actor-Critic (SAC)~\cite{sac} to update the model, while SAFE-GPT leverages a textual fragment representation (SAFE) to generate molecules token-by-token and uses PPO to update the model. Unlike the vocabulary-based approach discussed below, which mines property-specific substructures (rationales) from labeled data, these methods use a learned policy to assemble fragments from general-purpose libraries.

\textbf{Vocabulary-Based Methods.} Vocabulary-based generation grounds the optimization process by recalling known high-value substructures. To circumvent the sparsity of high-reward molecules, RationaleRL~\cite{rationale_RL} extracts property-specific ``rationales'' (active subgraphs) with MCTS guided by property predictors and trains a generator that expands them into complete molecules. The generator is then fine-tuned with reinforcement learning. By treating generation as the completion of these building blocks, the method amortizes the search cost. However, it suffers from vocabulary rigidity: exploration is bounded by the combinatorial closure of the pre-extracted fragments.

\textbf{Conditional Encoder-Decoder Models.}
Distinct from stepwise graph construction, these models treat scaffold decoration as a translation task: mapping a starting scaffold to a completed molecule. SyntaLinker~\cite{syntalinker} applies this paradigm to 1D sequences, utilizing a Transformer to ``translate'' fragment pairs into linked molecules. LibINVENT~\cite{libinvent} likewise conditions on a scaffold: an encoder--decoder RNN generates its R-groups, and its reinforcement learning stage is anchored to a pretrained prior. In the graph domain, DrugEx v3~\cite{drugex_v3} employs an encoder-decoder Graph Transformer trained via a multi-objective REINFORCE algorithm. It explicitly conditions on the scaffold input, learning a policy to generate complementary structures. To manage the high variance typical of policy gradient updates, DrugEx v3 relies on a Pareto-based ranking strategy and an explicit ``exploration network'' (a fixed, pretrained copy of the policy) to force diverse sampling. However, unlike our approach which centers rewards relative to the specific scaffold instance (GRPO), DrugEx v3's updates utilize a global reward formulation. Consequently, to prevent mode collapse on difficult scaffolds, it must rely on external exploration heuristics rather than an intrinsic variance-reduction mechanism.

\textbf{Search-Based Imitation Learning.}
A hybrid approach involves training a policy to imitate the trajectory of a search algorithm. GraphXForm~\cite{graphxform} employs a Self-Improvement Learning (SIL) loop, combining the Deep Cross-Entropy Method (CEM) with a sequence decoding algorithm ``Take a Step and Reconsider'' (TASAR)~\cite{tasar}. In this paradigm, the model generates candidate sequences, evaluates them with the oracle, and selects the highest-scoring trajectories as pseudo-labels for supervised fine-tuning. While this stabilizes training without value approximation, it remains computationally expensive during the training phase. It is a known phenomenon that supervised (imitation) learning generalizes less effectively than reinforcement learning~\cite{generalize, gap, sft}.

\textbf{Our Approach:} \textbf{AMORTIX} incorporates the structural inductive biases of Graph Transformers while utilizing atom-level stepwise graph construction, and amortizes the optimization cost effectively into a learned policy. By using group-based statistics as a dynamic baseline, GRPO stabilizes the learning of constructive design rules across diverse starting points.

\subsection{Structurally-Constrained Generation Tasks}
\label{app:constrained_tasks}
\begin{figure}[H]
    \centering
    \includegraphics[width=0.95\linewidth]{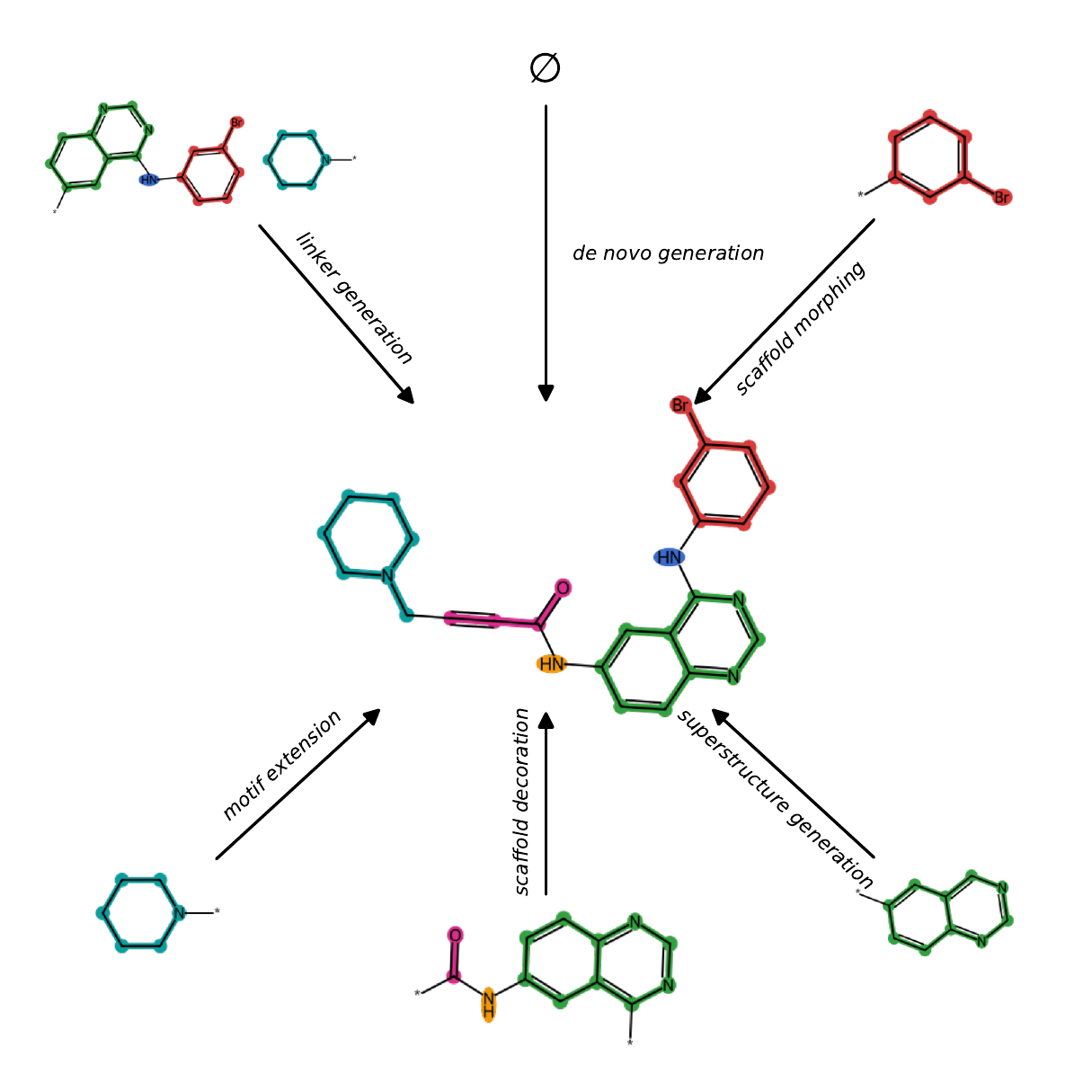}
    \caption{Structurally-constrained molecular design tasks. Each task corresponds to a different choice of initial
    subgraph $s_0$ (outer molecules), to which atoms and bonds are added to construct the target molecule (center).}
    \label{fig:constrained_tasks}
\end{figure}

Structurally-constrained molecular design encompasses a range of objectives
where specific substructures must be preserved~\cite{leadopt, safe};
Figure~\ref{fig:constrained_tasks} illustrates the variants most commonly
considered in the literature:
\begin{itemize}
    \setlength{\itemsep}{2pt}
    \setlength{\parskip}{0pt}
    \setlength{\topsep}{2pt}
    \item \textbf{Scaffold decoration:} the fixed core structure (ring systems
    and linkers) is preserved while the agent attaches functional groups or
    side chains.
    \item \textbf{Motif extension:} a small structural fragment is extended
    into a larger, complete molecular graph.
    \item \textbf{Superstructure generation:} the agent generates a molecule
    containing a specific input substructure.
    \item \textbf{Linker design:} a bridge is generated to connect two or more
    disjoint fragments.
    \item \textbf{Lead optimization:} a high-affinity molecule is modified to
    improve desired properties.
    \item \textbf{Scaffold morphing:} the core scaffold is modified (e.g., via
    ring expansion) while retaining specific substituent patterns.
\end{itemize}
In our framework, these tasks are treated uniformly: generation initiates from
a non-empty subgraph $s_0$, and the policy sequentially adds atoms and bonds.
By setting $s_0$ to a single atom, the formulation also encompasses
\textit{de novo} generation.

\section{AMORTIX Architecture}\label{model_architecture}
\begin{figure}[t]
\centering
\includegraphics[width=0.9\textwidth]{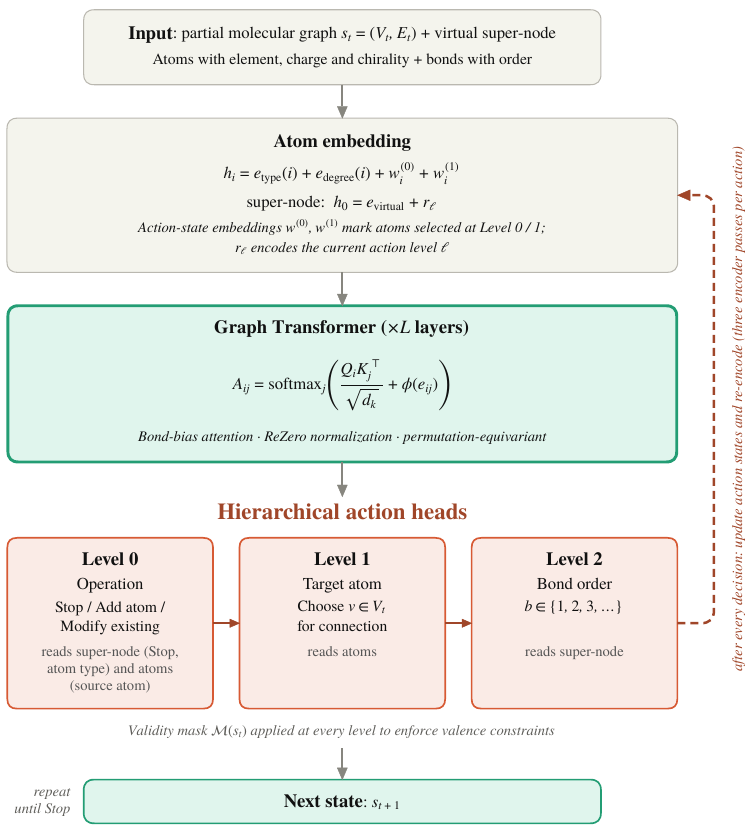}
\caption{AMORTIX policy architecture. A partial molecular graph $s_t$ and a
virtual super-node are embedded and processed by an $L$-layer Graph Transformer
with bond-order bias in the attention scores. A composite action is decoded in
three sequential steps (Level 0: operation; Level 1: target atom; Level 2: bond
order), each with its own encoder pass: after every decision, the action-state
embeddings $w^{(0)}, w^{(1)}$ and the action-level embedding $r_\ell$ of the
super-node are updated and the graph is re-encoded. A validity mask
$\mathcal{M}(s_t)$ removes valence-violating actions at every level, so every
generated molecule is chemically valid by construction.}
\label{fig:architecture}
\end{figure}

Figure~\ref{fig:architecture} provides a high-level overview of the AMORTIX 
policy. The remainder of this appendix details each component: the action 
space, the atom vocabulary, the policy 
architecture, and the pretraining procedure.

\subsection{Action Space}
We model molecular generation as a sequential Markov Decision Process. A state $s_t$ represents the intermediate partial molecular graph at step $t$. This graph consists of a set of atoms $V_t$ (with properties such as element type, charge, etc.) and bonds $E_t$.

The action space $\mathcal{A}$ is decomposed hierarchically to account for the discrete nature of graph construction. To transition from state $s_t$ to state $s_{t+1}$, the policy samples a composite action that is decomposed into three sequential sub-levels. This decomposition reduces the combinatorial complexity of the action space:

\begin{enumerate}
    \item \textbf{Action Level 0 (Operation Selection):} The agent first decides the nature of the modification. It produces a distribution over three categories of moves:
    \begin{itemize}
        \item \textit{Termination:} The \textsc{Stop} token, ending the generation process.
        \item \textit{Add Atom:} Selecting a new element type $T \in \Sigma$ (e.g., C, N, O, ...) from the vocabulary to add to the graph. For all our experiments, the vocabulary size is 23 (see Appendix \ref{vocab} for details).
        \item \textit{Modify Existing:} Selecting an existing atom $u \in V_t$ to serve as the source of a new bond.
    \end{itemize}
    \item \textbf{Action Level 1 (Target Selection):} Conditional on the first selection, the agent selects a target atom $v \in V_t$. If a new atom type was selected in Level 0, this step determines the anchor point $v$ on the existing graph to which the new atom connects. If an existing atom $u$ was selected in Level 0, $v$ represents the destination of a new bond starting from $u$.
    \item \textbf{Action Level 2 (Bond Specification):} Finally, the agent determines the bond order $b \in \{1,2,3,4,5,6\}$ for the edge connecting the two atoms identified in the previous steps. The range follows RDKit's \texttt{BondType} enum (single, double, triple, quadruple, quintuple, hextuple). Bond orders 4--6 are retained for completeness but are not realized in any generated molecule in our experiments; in practice the policy assigns mass to $b \in \{1,2,3\}$.
\end{enumerate}

To ensure chemical validity, we employ a validity mask $\mathcal{M}(s_t)$. At each action level, probabilities of actions that would violate valence constraints (e.g., exceeding the maximum bond capacity of an atom) are masked to zero. This guarantees that every sampled trajectory remains within the subspace of chemically valid graphs.

\subsection{Atom Vocabulary}\label{vocab}
The generative model uses a fixed vocabulary of 23 atomic tokens, representing specific combinations of element, formal charge, and chirality. The allowed tokens are:
\begin{itemize}
    \setlength\itemsep{0.1em}
\item \textbf{Carbon:} \texttt{C}, \texttt{C$^+$}, \texttt{C$^-$}, \texttt{C$@$}, \texttt{C$@@$}
\item \textbf{Nitrogen:} \texttt{N}, \texttt{N$^+$}, \texttt{N$^-$}
\item \textbf{Oxygen:} \texttt{O}, \texttt{O$^+$}, \texttt{O$^-$}
\item \textbf{Phosphorus:} \texttt{P}, \texttt{P$^+$}, \texttt{P$^-$}
\item \textbf{Sulfur:} \texttt{S}, \texttt{S$^+$}, \texttt{S$^-$}, \texttt{S$@$}, \texttt{S$@@$}
\item \textbf{Halogens:} \texttt{F}, \texttt{Cl}, \texttt{Br}, \texttt{I}
\end{itemize}
Note: \texttt{@} and \texttt{@@} denote tetrahedral chirality attributes.

\subsection{Policy Architecture}
We parameterize the policy $\pi_\theta$ using a decoder-only Graph Transformer that processes the molecule as an unordered set of atoms. To capture global context and aggregate sequence-level information, the graph is augmented with a virtual ``super-node'' $v_{virtual}$ connected to all atoms via special edges. This virtual node aggregates 
global graph context and is later used by the Level 0 action head to 
predict graph-level decisions (Stop, or new atom type).

\paragraph{Input Representation.} 
The input to the model is the set of atoms in the current partial graph. To condition the policy on the sequential sub-actions (e.g., informing Level 1 which atom was picked in Level 0), we augment the standard atom features with dynamic state embeddings. The input embedding $h_i$ for the $i$-th atom is the sum of:
\begin{itemize}
    \item \textbf{Atom Type Embedding:} A learnable vector representing the element type.
    \item \textbf{Degree Embedding:} Encoding the current number of bonds connected to the atom.
    \item \textbf{Action State Embeddings:} Two binary embedding vectors $w^{(0)}_i$ and $w^{(1)}_i$ which indicate whether atom $i$ was selected at Action Level 0 or Level 1, respectively. This allows the transformer to ``focus'' on the active atoms during the multi-step decision process.
\end{itemize}

The embedding of the virtual super-node is the sum of its own learnable vector $e_{\mathrm{virtual}}$
and a learnable embedding $r_\ell$ of the current action level $\ell$. After each level's
decision, the chosen information is fed back by updating the action-level and
action-state embeddings before the next pass.

\paragraph{Graph Transformer Layers.}
These embeddings are processed through $L$ layers of Multi-Head Self-Attention (MHSA) with ReZero normalization~\cite{rezero}. We do not use positional encodings, preserving permutation equivariance. Structural topology is injected directly into the attention mechanism via bias terms. The attention score $A_{ij}$ between atoms $i$ and $j$ is computed as:
\begin{equation}
    A_{ij} = \text{softmax}\left(\frac{Q_i K_j^T}{\sqrt{d_k}} + \phi(e_{ij})\right)
\end{equation}

where $\phi(e_{ij})$ is a learnable scalar bias derived from the bond order $e_{ij}$ connecting the atoms. This allows the model to attend to topological neighbors differently than distant atoms.

\paragraph{Action Heads.}
The three levels are decoded in three separate forward passes
through the full encoder: after each decision the action-level and action-state
embeddings are updated and the graph is re-encoded. The Level 0 head projects
the virtual node representation to logits for \textsc{Stop} and the new atom
types, and separately projects each atom representation to a logit for selecting
an existing atom as the first atom of a new bond. The Level 1 head projects each
atom representation to a logit for the second atom, and the Level 2 head
projects the virtual node representation to the bond-order logits.

\subsection{Pretraining}
Before optimizing for specific molecular properties, we initialize the policy to capture the fundamental syntax of stable chemistry. We utilize the ChEMBL database~\cite{chembl} (version 35) as a source of chemically valid structures. Following the pretraining protocol of GraphXForm~\cite{graphxform}, we filtered the database to include only molecules composed of atoms within our defined alphabet. This resulted in a training set of approximately 1.5 million molecules and a validation set of roughly 70,000 molecules.

Since the policy $\pi_\theta$ operates sequentially, we transform each molecule
$G \in \mathcal{D}_{pre}$ into a construction trajectory $\tau^*$. We compute
the canonical SMILES, kekulize the molecule, and emit atoms in canonical index
order, with atom~0 as the root. Each subsequent atom $i$ yields one \textit{Add
Atom} action that attaches it, with the appropriate bond order, to its
lowest-indexed neighbor $j<i$, followed by \textit{Modify Existing} actions for
its remaining neighbors of lower index (ring closures). Atom tokens carry
element, formal charge and chirality, so charge and stereochemistry are
preserved. Every trajectory is checked to reproduce the canonical SMILES of its
molecule, which holds for almost all training molecules.

Because bonds are only added between a new atom and existing neighbors, the
accumulated bond order of an atom never exceeds its final bond order, so the
validity mask never blocks a ground-truth trajectory. Every intermediate graph is valence-valid, though not
necessarily stable (for instance, a partial aromatic ring). As in
autoregressive language modeling, where prefixes of sentences are not sentences,
the policy learns a distribution over construction steps whose endpoints are
real compounds. The canonical decomposition is a modeling choice: it fixes one
traversal order, and ring closures are always emitted immediately onto the
most recently added atom. Reinforcement learning is on-policy and not anchored
to this prior (Section~\ref{sec:grpo}), but we discuss the potential bias in the
Limitations.

We then train the policy by supervised next-action prediction, as in
GraphXForm~\cite{graphxform}. In each of the 1.5 million training steps we
sample a batch of $M=512$ state--action pairs $(s_t^*, a_t^*)$ uniformly at
random from all steps of the trajectories of the pretraining molecules (so that
larger molecules contribute more examples), that is, partial graphs together with
the ground-truth next sub-action, and minimize the
cross-entropy between the predicted action distribution and $a_t^*$:
\begin{equation}
    \mathcal{L}_{\text{MLE}}(\theta) = - \frac{1}{M} \sum_{b=1}^{M} \log \pi_\theta\!\left(a^*_{t_b} \mid s^*_{t_b}\right),
\end{equation}
where $s_t^*$ is the partial graph resulting from the ground-truth prefix
$a^*_{0:t-1}$ and $a_t^*$ is the ground-truth sub-action at its current level
(Level 0, 1 or 2), each sub-action being a separate example. Pretraining used a
dropout rate of $0.1$.

This step ensures that the policy generates chemically plausible structures before optimization begins, thus making the subsequent RL fine-tuning phase tractable.

\subsection{Fine-tuning}\label{app:finetuning}
During the RL fine-tuning phase, the policy is updated according to Algorithm~\ref{alg:AMORTIX_inner}.

\begin{figure}[H]
    \centering
    \includegraphics[scale=0.7]{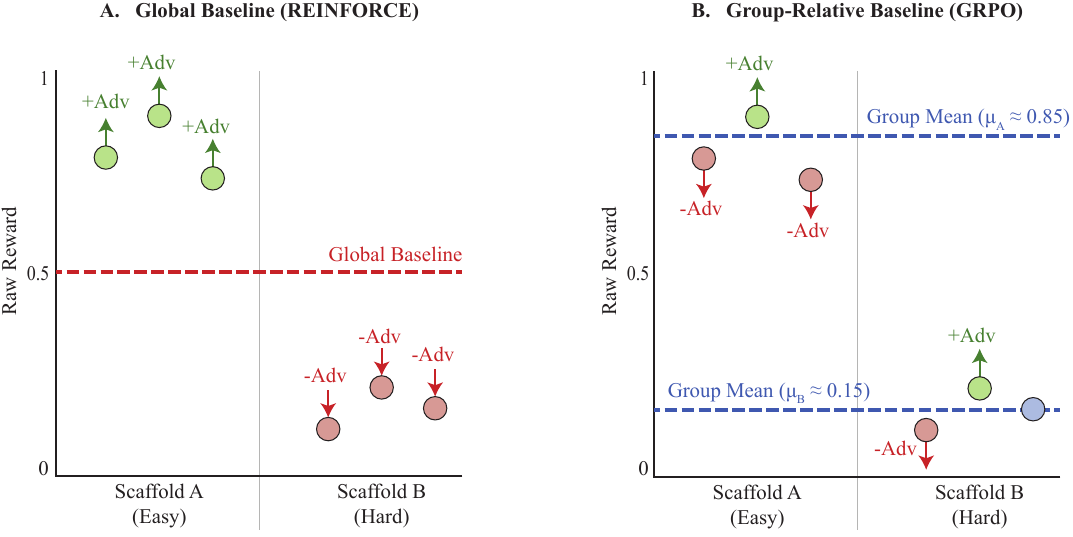}
    \caption{Schematic comparison of advantage estimation strategies (illustrative,
not measured data). \textbf{(A) Global baseline (REINFORCE):} a single mean over
all starting structures mixes scaffold difficulty with policy quality, which
inflates the variance of the advantages. \textbf{(B) Group-relative baseline
(GRPO):} per-structure means ($\mu_A, \mu_B$) make each advantage reflect only
the choices made on that structure, stabilizing the learning signal across easy
and hard structures. The measured counterpart on three tasks is shown in Figure~\ref{fig:variance}.}
    \label{fig:grpo}
\end{figure}

The heterogeneous difficulty of starting structures produces reward signals on incomparable scales, motivating the use of GRPO in this setting (Figure~\ref{fig:grpo}).
\begin{algorithm}[h]
\caption{AMORTIX Training Loop}
\label{alg:AMORTIX_inner}
\begin{algorithmic}[1] 
    \STATE {\bfseries Input:} Set of starting structures $\mathcal{D}$, Reward oracle $R$, Initial policy $\pi_\theta$
    \STATE {\bfseries Hyperparameters:} Starting structures per batch $B$, group size $G$, learning rate $\alpha$
    \REPEAT
    \STATE Sample batch of starting structures $\{S_1, \dots, S_B\} \sim \mathcal{D}$
    \STATE Initialize gradient accumulator $\Delta \theta \leftarrow 0$
    \FOR{$i=1$ {\bfseries to} $B$}
        \STATE Generate group $\mathcal{O}_i = \{O_{i,1}, \dots, O_{i,G}\}$ using IID MC sampling from $\pi_\theta(\cdot|S_i)$ 
        \STATE Compute rewards $r_{i,j} \leftarrow R(O_{i,j})$ for all $j \in \{1, \dots, G\}$
        \STATE Compute group mean $\mu_i \leftarrow \frac{1}{G} \sum_{k=1}^G r_{i,k}$
        
        \FOR{$j=1$ {\bfseries to} $G$}
            \STATE $A_{i,j} \leftarrow r_{i,j} - \mu_i$ 
            \STATE \COMMENT{Accumulate gradients over generation steps $t$}
            \STATE $\Delta \theta \leftarrow \Delta \theta + A_{i,j} \sum_{t=1}^{T_{i,j}} \nabla_\theta \log \pi_\theta(a_t|s_{<t})$
        \ENDFOR
    \ENDFOR
    \STATE Update parameters with Adam along the ascent direction $\frac{1}{B \cdot G} \Delta \theta$
    \UNTIL{convergence}
\end{algorithmic}
\end{algorithm}

\paragraph{Entropy Regularization.}
To prevent premature convergence and encourage exploration, the objective may be augmented with an entropy regularization term. We calculate the entropy $H(\pi_\theta(\cdot | s_t))$ of the policy distribution at each step $t$. To ensure numerical stability and validity, we strictly compute this over valid actions by filtering out tokens masked by the valence constraints:
\begin{equation}
    H(\pi_\theta(\cdot | s_t)) = - \sum_{k \in \mathcal{A}_{valid}} \pi_\theta(a_k | s_t) \log \pi_\theta(a_k | s_t)
\end{equation}
where $\mathcal{A}_{valid}$ is the set of chemically valid actions at state $s_t$.

The final gradient update is performed by maximizing the joint objective $\mathcal{J}_{total} = \mathcal{J}_{RL} + \beta H$. Since sampling trajectories is inexpensive compared to parameter updates, we perform a single gradient update per batch using the following estimator:
\begin{equation}
\nabla_\theta \mathcal{J}(\theta) \approx \frac{1}{BG} \sum_{i=1}^B \sum_{j=1}^G \left[ A_{i,j} \sum_{t=1}^{T_{i,j}} \nabla_\theta \log \pi_\theta(a_t | s_{<t}) + \beta \sum_{t=1}^{T_{i,j}} \nabla_\theta H(\pi_\theta(\cdot | s_{<t})) \right]
\end{equation}
Here, $A_{i,j}$ is the group-centered advantage for GRPO (for the REINFORCE runs
on PMO it is computed against a global moving-average baseline), and $\beta$ is
the entropy coefficient. While all results reported in this
paper use $\beta=0$ for every task, i.e., no per-task tuning of $\beta$, during the course of our experiments we did note that certain tasks benefit from small regularization values, e.g., $\beta \in \{0.001, 0.002\}$.

\section{Dataset Preparation and Scaffold Splitting}
\label{app:dataset_split}

\begin{figure}[t]
    \centering
    \includegraphics[width=\linewidth]{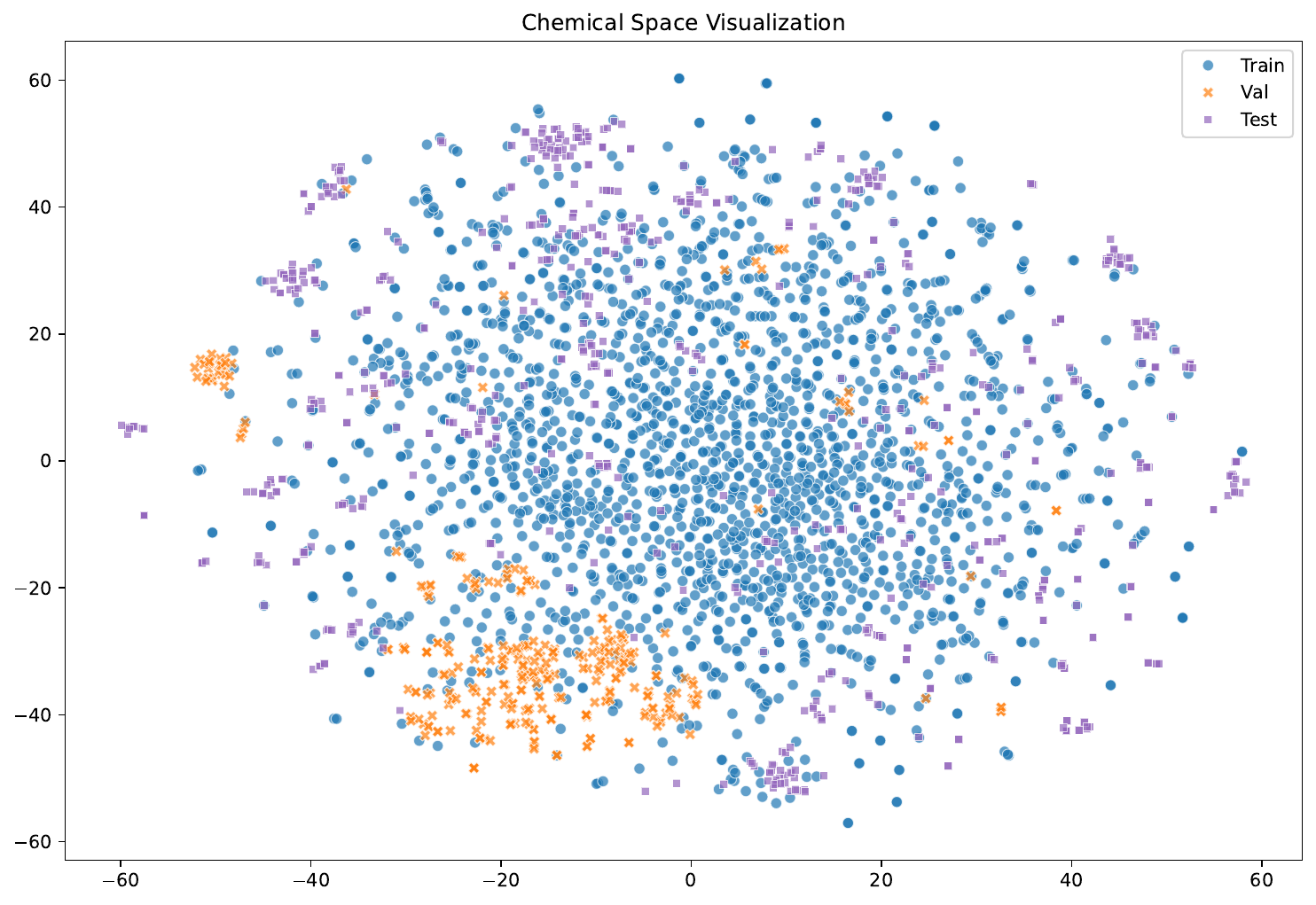}
    \caption{\textbf{Structural Generalization Split.} t-SNE visualization of the chemical space (Morgan Fingerprints) for Training, Validation, and Test scaffolds. The cluster-based splitting strategy ensures that Test scaffolds (purple) occupy distinct regions of chemical space compared to the Training set (blue) and Validation set (orange), enforcing a test of out-of-distribution generalization.}
    \label{fig:tsne}
\end{figure}

To evaluate the generalization capability of the model, we utilized the ZINC-250k dataset~\cite{zinc}. A rigorous cluster-based split was implemented to ensure that the test set consists of topologically distinct scaffolds rather than local structural variations.

We performed Murcko scaffold decomposition using RDKit~\cite{rdkit} and clustered the resulting scaffolds using the Butina algorithm~\cite{butina} on Morgan fingerprints (radius 2, 2048 bits) with a Tanimoto similarity cutoff of 0.4. Entire clusters were assigned to training, validation, or testing, reserving 500 structurally distinct scaffolds for the test set.

Figure~\ref{fig:tsne} visualizes the chemical space of the resulting splits. The clear separation between training (blue) and test (purple) distributions confirms that the model must learn to generalize optimization rules to unseen regions of chemical space rather than memorizing training examples.

\section{PMO Benchmark}\label{pmo}

The PMO benchmark consists of \textit{de novo} tasks: each task has a single objective
and generation starts from scratch rather than from a given structure. It is therefore not the amortized setting this paper
targets, and we include it as a sanity check of the policy architecture. To
adhere to the PMO protocol of a $10{,}000$-call budget per task, we run AMORTIX
as an online optimizer: starting from the ChEMBL-pretrained policy, we fine-tune
it separately for each task with online RL. As we implement \textit{de novo} generation
by initializing every episode with a single carbon atom, there is no variation
in starting states. The group-relative advantage is therefore not needed, and we
use standard REINFORCE with a global baseline. Consequently, this benchmark does
not test GRPO.

The PMO benchmark evaluates sample efficiency using the Area Under the Curve (AUC) of the top-10 property scores, calculated via the trapezoidal rule at intervals of 100 oracle calls. This metric plots the average score of the top-10 molecules found so far ($y$-axis) against the number of oracle calls ($x$-axis) consumed during the online training trajectory. By integrating this curve, the metric penalizes methods that require a long warm-up period or extensive exploration to find their first high-scoring candidates. A high AUC indicates that the agent rapidly locates high-scoring molecules early in the learning process.

We compare AMORTIX against the highest-performing baselines reported in~\cite{genmol} and~\cite{syngbo},
which cover discrete diffusion (GenMol), synthesis-aware optimization (SynGBO),
retrieval-augmented generation (f-RAG), evolutionary and flow-based strategies
(Mol~GA, Genetic GFN), reinforcement learning (REINVENT) and Bayesian
optimization (GP BO). Baseline numbers are taken from these two sources: GenMol, whose results average
3 runs, from~\cite{genmol}, and all others from the compilation in~\cite{syngbo}, which averages 5 runs (3 for f-RAG) and takes Mol~GA and REINVENT from~\cite{genetic_gfn}. AMORTIX is averaged over 3 runs.

Pretraining data differ across methods: AMORTIX is pretrained on ChEMBL~35,
whereas GenMol and f-RAG build on SAFE-based models pretrained on a
billion-scale dataset~\cite{safe} (Appendix~\ref{app:pretraining}).

\paragraph{Results.}
Table~\ref{tab:results_comparison} reports the top-10 AUC on the 22 tasks.
AMORTIX reaches an aggregate score of $16.213$. This is above Genetic GFN
($16.078$), Mol~GA ($15.686$) and REINVENT ($15.003$) and below GenMol
($17.540$), SynGBO ($16.426$), GP~BO ($16.304$) and f-RAG ($16.301$). It attains
the best score, in one case tied with GP~BO, on two tasks (Celecoxib
Rediscovery and JNK3) and is among the top three on eleven. Given the
architecture-only setting, we regard this as competitive with dedicated
optimizers, but not better than the strongest ones.

\begin{table*}[t]
\caption{PMO benchmark (top-10 AUC, 22 tasks). AMORTIX is run as an online
optimizer with REINFORCE and a fixed entropy coefficient $\beta=0$ on every task
(mean $\pm$ SD over 3 runs). Baseline results are taken from their respective
publications (3 to 5 runs), as compiled in~\cite{syngbo} (Mol~GA and REINVENT
from~\cite{genetic_gfn}), except GenMol, which is taken from~\cite{genmol}.
All methods optimize online within a 10{,}000-call budget. The top three results
per task are colored
\textcolor[HTML]{0072B2}{\textbf{blue}} (1st),
\textcolor[HTML]{E69F00}{\textbf{orange}} (2nd) and
\textcolor[HTML]{009E73}{\textbf{green}} (3rd); tied scores share a color.}
\label{tab:results_comparison}
\vskip 0.1in
\begin{center}
\begin{small}
\begin{sc}
\resizebox{\textwidth}{!}{
\begin{tabular}{lcccccccc}
\toprule
\textbf{Task} & \textbf{AMORTIX} & \textbf{f-RAG} & \textbf{Genetic GFN} & \textbf{REINVENT} & \textbf{GenMol} & \textbf{SynGBO} & \textbf{GP BO} & \textbf{Mol~GA} \\
& \textbf{(Ours)} & & & & & & & \\
\midrule
Albuterol Similarity & \ranktwo{0.972} $\pm$ 0.004 & \rankone{0.977} $\pm$ 0.002 & 0.949 $\pm$ 0.010 & 0.881 $\pm$ 0.016 & 0.937 $\pm$ 0.010 & 0.947 $\pm$ 0.024 & \rankthree{0.964} $\pm$ 0.050 & 0.928 $\pm$ 0.015 \\
Amlodipine MPO & 0.666 $\pm$ 0.021 & \rankthree{0.749} $\pm$ 0.019 & \ranktwo{0.761} $\pm$ 0.019 & 0.644 $\pm$ 0.019 & \rankone{0.810} $\pm$ 0.012 & 0.670 $\pm$ 0.088 & 0.720 $\pm$ 0.061 & 0.740 $\pm$ 0.055 \\
Celecoxib Rediscovery & \rankone{0.860} $\pm$ 0.005 & 0.778 $\pm$ 0.007 & 0.802 $\pm$ 0.029 & 0.717 $\pm$ 0.027 & \rankthree{0.826} $\pm$ 0.018 & \ranktwo{0.856} $\pm$ 0.013 & \rankone{0.860} $\pm$ 0.002 & 0.629 $\pm$ 0.062 \\
Deco Hop & \ranktwo{0.944} $\pm$ 0.026 & \rankthree{0.936} $\pm$ 0.011 & 0.733 $\pm$ 0.109 & 0.662 $\pm$ 0.044 & \rankone{0.960} $\pm$ 0.010 & 0.831 $\pm$ 0.039 & 0.672 $\pm$ 0.118 & 0.656 $\pm$ 0.013 \\
DRD2 & \rankthree{0.991} $\pm$ 0.001 & \ranktwo{0.992} $\pm$ 0.000 & 0.974 $\pm$ 0.006 & 0.957 $\pm$ 0.007 & \rankone{0.995} $\pm$ 0.000 & 0.981 $\pm$ 0.010 & 0.902 $\pm$ 0.117 & 0.950 $\pm$ 0.004 \\
Fexofenadine MPO & 0.829 $\pm$ 0.027 & \ranktwo{0.856} $\pm$ 0.016 & \ranktwo{0.856} $\pm$ 0.039 & 0.781 $\pm$ 0.013 & \rankone{0.894} $\pm$ 0.028 & 0.833 $\pm$ 0.018 & 0.806 $\pm$ 0.006 & \rankthree{0.835} $\pm$ 0.012 \\
GSK3$\beta$ & \rankthree{0.931} $\pm$ 0.017 & \ranktwo{0.969} $\pm$ 0.003 & 0.881 $\pm$ 0.042 & 0.885 $\pm$ 0.031 & \rankone{0.986} $\pm$ 0.003 & 0.924 $\pm$ 0.027 & 0.877 $\pm$ 0.055 & 0.894 $\pm$ 0.025 \\
Isomers C7H8N2O2 & \rankthree{0.958} $\pm$ 0.048 & 0.955 $\pm$ 0.008 & \ranktwo{0.969} $\pm$ 0.003 & 0.942 $\pm$ 0.012 & 0.942 $\pm$ 0.004 & \rankone{0.975} $\pm$ 0.006 & 0.911 $\pm$ 0.031 & 0.926 $\pm$ 0.014 \\
Isomers C9H10N2O2PF2Cl & 0.795 $\pm$ 0.038 & 0.850 $\pm$ 0.005 & \rankone{0.897} $\pm$ 0.007 & 0.838 $\pm$ 0.030 & 0.833 $\pm$ 0.014 & \rankthree{0.875} $\pm$ 0.013 & 0.828 $\pm$ 0.126 & \ranktwo{0.894} $\pm$ 0.005 \\
JNK3 & \rankone{0.914} $\pm$ 0.037 & 0.904 $\pm$ 0.004 & 0.764 $\pm$ 0.069 & 0.782 $\pm$ 0.029 & \rankthree{0.906} $\pm$ 0.023 & \ranktwo{0.910} $\pm$ 0.021 & 0.785 $\pm$ 0.072 & 0.835 $\pm$ 0.040 \\
Median 1 & 0.360 $\pm$ 0.036 & 0.340 $\pm$ 0.007 & \rankthree{0.379} $\pm$ 0.010 & 0.363 $\pm$ 0.011 & \ranktwo{0.398} $\pm$ 0.000 & 0.357 $\pm$ 0.001 & \rankone{0.415} $\pm$ 0.001 & 0.329 $\pm$ 0.006 \\
Median 2 & 0.277 $\pm$ 0.024 & 0.323 $\pm$ 0.005 & 0.294 $\pm$ 0.007 & 0.281 $\pm$ 0.002 & \ranktwo{0.359} $\pm$ 0.004 & \rankthree{0.349} $\pm$ 0.001 & \rankone{0.408} $\pm$ 0.003 & 0.284 $\pm$ 0.035 \\
Mestranol Similarity & \ranktwo{0.974} $\pm$ 0.010 & 0.671 $\pm$ 0.021 & 0.708 $\pm$ 0.057 & 0.634 $\pm$ 0.042 & \rankone{0.982} $\pm$ 0.000 & 0.759 $\pm$ 0.023 & \rankthree{0.930} $\pm$ 0.106 & 0.762 $\pm$ 0.048 \\
Osimertinib MPO & 0.848 $\pm$ 0.011 & \ranktwo{0.866} $\pm$ 0.009 & \rankthree{0.860} $\pm$ 0.008 & 0.834 $\pm$ 0.010 & \rankone{0.876} $\pm$ 0.008 & 0.856 $\pm$ 0.024 & 0.833 $\pm$ 0.011 & 0.853 $\pm$ 0.005 \\
Perindopril MPO & 0.520 $\pm$ 0.007 & \rankthree{0.681} $\pm$ 0.017 & 0.595 $\pm$ 0.014 & 0.535 $\pm$ 0.015 & \ranktwo{0.718} $\pm$ 0.012 & \rankone{0.774} $\pm$ 0.006 & 0.651 $\pm$ 0.030 & 0.610 $\pm$ 0.038 \\
QED & \rankthree{0.941} $\pm$ 0.001 & 0.939 $\pm$ 0.001 & \ranktwo{0.942} $\pm$ 0.000 & \rankthree{0.941} $\pm$ 0.000 & \ranktwo{0.942} $\pm$ 0.000 & 0.940 $\pm$ 0.002 & \rankone{0.947} $\pm$ 0.000 & \rankthree{0.941} $\pm$ 0.001 \\
Ranolazine MPO & \ranktwo{0.833} $\pm$ 0.012 & 0.820 $\pm$ 0.016 & 0.819 $\pm$ 0.018 & 0.770 $\pm$ 0.005 & 0.821 $\pm$ 0.011 & \rankone{0.839} $\pm$ 0.016 & 0.810 $\pm$ 0.011 & \rankthree{0.830} $\pm$ 0.010 \\
Scaffold Hop & \rankthree{0.588} $\pm$ 0.030 & 0.576 $\pm$ 0.014 & \ranktwo{0.615} $\pm$ 0.100 & 0.551 $\pm$ 0.024 & \rankone{0.628} $\pm$ 0.008 & 0.541 $\pm$ 0.008 & 0.529 $\pm$ 0.020 & 0.568 $\pm$ 0.017 \\
Sitagliptin MPO & 0.513 $\pm$ 0.074 & \rankthree{0.601} $\pm$ 0.011 & \ranktwo{0.634} $\pm$ 0.039 & 0.470 $\pm$ 0.041 & 0.584 $\pm$ 0.034 & 0.454 $\pm$ 0.074 & 0.474 $\pm$ 0.085 & \rankone{0.677} $\pm$ 0.055 \\
Thiothixene Rediscovery & 0.513 $\pm$ 0.058 & 0.584 $\pm$ 0.009 & 0.583 $\pm$ 0.034 & 0.544 $\pm$ 0.026 & \ranktwo{0.692} $\pm$ 0.123 & \rankthree{0.647} $\pm$ 0.003 & \rankone{0.727} $\pm$ 0.089 & 0.544 $\pm$ 0.067 \\
Troglitazone Rediscovery & 0.504 $\pm$ 0.023 & 0.448 $\pm$ 0.017 & 0.511 $\pm$ 0.054 & 0.458 $\pm$ 0.018 & \rankone{0.867} $\pm$ 0.022 & \rankthree{0.579} $\pm$ 0.002 & \ranktwo{0.756} $\pm$ 0.141 & 0.487 $\pm$ 0.024 \\
Zaleplon MPO & 0.482 $\pm$ 0.011 & 0.486 $\pm$ 0.004 & \ranktwo{0.552} $\pm$ 0.033 & \rankthree{0.533} $\pm$ 0.009 & \rankone{0.584} $\pm$ 0.011 & 0.529 $\pm$ 0.017 & 0.499 $\pm$ 0.025 & 0.514 $\pm$ 0.033 \\
\midrule
\textbf{Sum} & 16.213 & 16.301 & 16.078 & 15.003 & \rankone{17.540} & \ranktwo{16.426} & \rankthree{16.304} & 15.686 \\
\bottomrule
\end{tabular}
}
\end{sc}
\end{small}
\end{center}
\vskip -0.1in
\end{table*}

\paragraph{Task Exclusion.}
We evaluate on 22 of the 23 PMO tasks and exclude Valsartan SMARTS, following
\cite{syngbo}. The oracle returns zero unless the molecule contains the
substructure \texttt{CN(C=O)Cc1ccc(c2ccccc2)cc1}, so optimizers receive no
signal until they happen to propose it. This makes the task a test of
constraint satisfaction rather than of optimization efficiency.

\section{Amortization Analysis, Runtime, and Hardware Details}
\label{app:oracle_analysis}

A defining characteristic of amortized optimization is the decoupling of expensive oracle evaluations (e.g., docking simulations) from the inference phase. By shifting the computational burden to a one-time training period, the marginal cost of generating optimized molecules becomes negligible. In Table~\ref{tab:oracle_calls}, we contrast the fixed training budget of amortized policies against the linear cost scaling of instance-specific optimizers. For every method, an oracle call is counted only for a molecule that has not been scored before in the same run: for amortized methods, within one training run; for instance optimizers, within the search for one scaffold. Repeated molecules are not charged.

\begin{table}[h]
    \caption{Oracle consumption on the Kinase MPO task. For amortized models, values denote the cumulative calls allowed while training the policy. For instance optimizers, values denote the number of oracle calls until the mean best-so-far score over test scaffolds converges (Appendix~\ref{app:oracle_analysis}). AMORTIX becomes cheaper in cumulative cost than Mol~GA after $\approx 100$ test scaffolds and GenMol after $\approx 150$ test scaffolds; see Figure~\ref{fig:oracle_cost}. Within a run, only oracle calls for previously unseen molecules are counted.}
    \label{tab:oracle_calls}
    \begin{center}
    \begin{small}
    \begin{sc}
    \begin{tabular}{llcc}
        \toprule
        Type & Method & Training Cost & Inference Cost \\
         & & (Fixed Total) & (Per Instance) \\
        \midrule
        \multirow{4}{*}{Amortized} & \textbf{AMORTIX} & \textbf{$\approx$ 50{,}000} & \textbf{0} \\
        & LibINVENT & $\approx$ 50{,}000 & 0 \\
         & GraphXForm & $\approx$ 335{,}000 & 0 \\
         & DrugEx v3 & $\approx$ 640{,}000 & 0 \\
        \midrule
        \multirow{2}{*}{Instance} & Mol~GA & N/A & $\approx$ 500 \\
         & GenMol & N/A & $\approx$ 330 \\
        \bottomrule
    \end{tabular}
    \end{sc}
    \end{small}
    \end{center}
\end{table}

\begin{figure}[h!]
    \centering
    \includegraphics[width=0.7\linewidth]{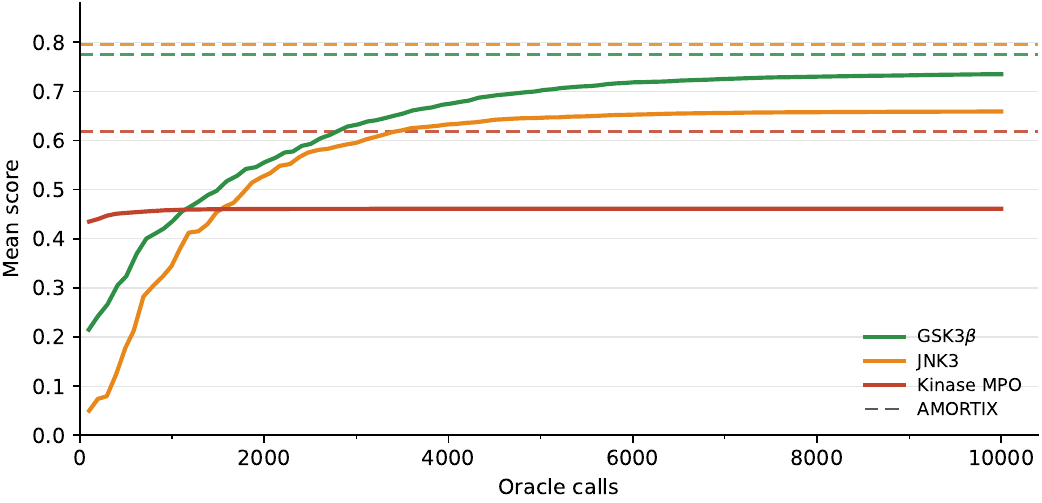}
    \caption{\textbf{Mol~GA convergence across objectives under hard scaffold constraints.}
Mean best score found so far as a function of cumulative oracle calls, averaged over all test scaffolds
(500 per split, 3 splits), for GSK3$\beta$, JNK3, and Kinase MPO.
Mol~GA was run for a fixed budget of 10{,}000 oracle calls per scaffold with no early stopping;
only scaffold-preserving, unique molecules count as oracle calls.
Dashed lines show AMORTIX's final scores from Table~\ref{tab:main_performance}.
Curves reach within $\approx$0.01 of their final value after roughly 500 oracle calls for Kinase MPO,
5{,}000 for JNK3, and 7{,}000 for GSK3$\beta$, which continues to climb slowly until the end of the budget.}
\label{fig:molga_convergence}
\end{figure}

\begin{figure}[h!]
    \centering
    \includegraphics[width=0.7\linewidth]{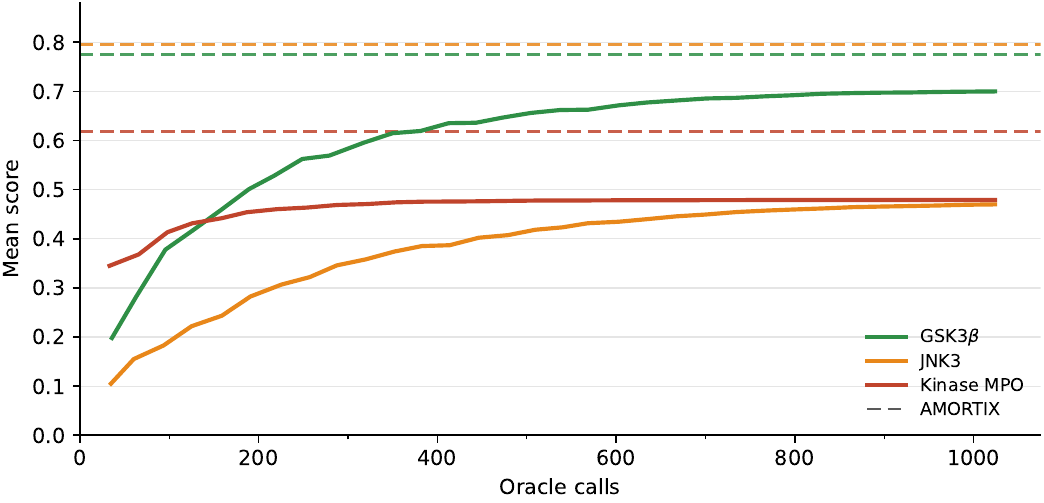}
    \caption{\textbf{GenMol convergence across objectives under hard scaffold constraints.}
Mean best score found so far as a function of cumulative oracle calls, averaged over all test scaffolds
(500 per split, 3 splits), for GSK3$\beta$, JNK3, and Kinase MPO.
GenMol was run for 32 generations of 32 samples ($\approx$1{,}000 calls per scaffold).
Dashed lines show AMORTIX's final scores from Table~\ref{tab:main_performance}.
Curves reach within $\approx$0.01 of their final value after roughly 330 oracle
calls for Kinase MPO, and after roughly 800 (GSK3$\beta$) and 900 (JNK3) calls.}
    \label{fig:genmol_convergence}
\end{figure}

\paragraph{Convergence criterion.}
For the amortization analysis, we define the convergence point of an
instance optimizer as the smallest number of oracle calls at which the mean best-so-far score, averaged over all test scaffolds, is within $0.01$ of its final value. Only the calls up to this point are charged to the instance optimizer. For Mol~GA and GenMol, an oracle call is counted only for a
scaffold-preserving molecule that has not been scored before; candidates that
violate the scaffold constraint and duplicate candidates are not charged.

We define the amortization break-even point as the threshold where the 
cumulative cost of instance optimization exceeds AMORTIX's fixed training 
budget; Figure~\ref{fig:oracle_cost} (main text) visualizes this break-even 
relative to Mol~GA and GenMol. The training costs of other amortized methods 
(GraphXForm: 335k calls; DrugEx v3: 640k calls) for the Kinase MPO task are reported in  Table~\ref{tab:oracle_calls} for completeness. Figures \ref{fig:molga_convergence} and \ref{fig:genmol_convergence} visualize the rate of convergence of the mean score against oracle calls for the instance optimizers (Mol~GA and GenMol) on all three goal-directed scaffold decoration tasks.

\paragraph{Runtime.} Beyond oracle counts, wall-clock latency presents a significant bottleneck.
At inference, GenMol needs
approximately 20--25 minutes to complete 32 optimization generations for one scaffold, whereas we measured that
AMORTIX takes on average 0.05--0.27 seconds to generate one completion per scaffold (batch size 1, greedy decoding, single GPU, excluding oracle scoring), growing with the number of atoms added at roughly 0.01 seconds per atom. For training, one AMORTIX run (500 epochs) takes about 4--5 hours of wall-clock time, for all variants. The 23-objective ablation consists of $3 \text{ variants} \times 3 \text{ scaffold splits} \times 23 \text{ objectives} = 207$ such runs, and the de novo variant adds 9
more, so it consumed about 1{,}000 GPU-hours in total for training.

\paragraph{Hardware.} All runs were conducted using one NVIDIA H100 GPU (96 GB HBM2e memory), an Intel Xeon 8468 Sapphire Rapids CPU (2.1 GHz, 24 usable cores), and up to 131 GB of usable system RAM per job. However, per run, actual peak GPU memory usage typically stayed below 16 GB and RAM usage below 32 GB.

\section{Hyperparameters and Configuration}
\label{app:hyperparams}

\subsection{Pretraining Datasets}\label{app:pretraining}
Table~\ref{tab:pretraining} lists the data on which each compared method was
pretrained, in the goal-directed scaffold decoration experiments (Section~\ref{subsec:kinase}) and on the
PMO benchmark (Appendix~\ref{pmo}). Mol~GA, GP~BO and SynGBO use no pretrained model and are not
listed; SynGBO instead searches over 196{,}907 purchasable Enamine building blocks
with 91 reaction templates~\cite{syngbo}. ZINC-250k serves in our scaffold
experiments only as the source of scaffolds (Appendix~\ref{app:dataset_split}).

\begin{table}[h]
\caption{Pretraining data of each compared method. S: goal-directed scaffold
decoration (Section~\ref{subsec:kinase}); P: PMO benchmark (Appendix~\ref{pmo}).}
\label{tab:pretraining}
\begin{center}
\footnotesize
\setlength{\tabcolsep}{4pt}
\begin{tabular}{L{2.2cm} L{1.0cm} L{5.4cm} L{3.6cm}}
\toprule
Method & Used in & Pretraining data & Training size \\
\midrule
AMORTIX & S, P & ChEMBL 35~\cite{chembl} & $\approx$1.5M molecules \\
GraphXForm~\cite{graphxform} & S & ChEMBL 35 (same as AMORTIX) & $\approx$1.5M molecules \\
LibINVENT~\cite{libinvent,reinvent4} & S & ChEMBL 27, sliced into scaffold--decoration pairs & 23.1M pairs \\
DrugEx v3~\cite{drugex_v3} & S & ChEMBL 27 & $\approx$1.7M molecules \\
GenMol~\cite{genmol} & S (V2), P (V1) & SAFE dataset~\cite{safe} (ZINC + UniChem) & 945M molecules \\
f-RAG~\cite{frag} & P & SAFE-GPT~\cite{safe} (ZINC + UniChem); ZINC-250k for the injection module & 1.1B molecules (SAFE-GPT) \\
Genetic GFN~\cite{genetic_gfn} & P & ZINC-250k (PMO protocol~\cite{pmo}) & $\approx$250k molecules \\
REINVENT~\cite{pmo, genetic_gfn} & P & ZINC-250k (PMO protocol) & $\approx$250k molecules \\
\bottomrule
\end{tabular}
\end{center}
\end{table}

\subsection{AMORTIX Implementation Details}
\label{app:AMORTIX-params}

This appendix subsection details the network architecture, optimization settings, and hyperparameters used for the AMORTIX experiments, as well as the GraphXForm baseline.

\subsubsection{Network Architecture and Optimization}
Table~\ref{tab:global_params} summarizes the global hyperparameters used for AMORTIX and GraphXForm (the RL parameters are only relevant for the former). For all AMORTIX and GraphXForm experiments, we restrict the maximum number of non-hydrogen atoms in a generated molecule to 50.

\begin{table}[h!]
    \centering
    \caption{Global Network Architecture and Optimization Hyperparameters.}
    \label{tab:global_params}
    \begin{tabular}{l l}
        \toprule
        \textbf{Parameter} & \textbf{Value} \\
        \midrule
        \multicolumn{2}{c}{\textit{Network Architecture}} \\
        Latent Dimension & 512 \\
        Number of Transformer Blocks & 10 \\
        Number of Attention Heads & 16 \\
        \midrule
        \multicolumn{2}{c}{\textit{Optimization (ADAM)}} \\
        Learning Rate & $1 \times 10^{-4}$ \\
        Weight Decay & 0 \\
        Gradient Clipping (L2-norm) & 1.0 \\
        LR Schedule Decay Factor & 1.0 (Constant) \\
        \midrule
        \multicolumn{2}{c}{\textit{Reinforcement Learning (GRPO)}} \\
        GRPO Updates per Epoch & 1 \\
        Entropy Coefficient $\beta$ & 0.0 \\
        \bottomrule
    \end{tabular}
\end{table}

\subsubsection{Experiment-Specific Configurations}
Table~\ref{tab:experiment_configs} outlines the differences between AMORTIX-DeNovo, AMORTIX-REINFORCE, AMORTIX-GRPO, AMORTIX-PPO, and the GraphXForm baseline. Each configuration is run independently for all three Goal-Directed Scaffold Decoration objectives (GSK3$\beta$, JNK3, and Kinase MPO) with reward functions defined in Section~\ref{subsec:kinase} (Equation~\ref{eq:kinase} and the corresponding single-objective forms). Hyperparameters are otherwise identical across the three runs.

AMORTIX-DeNovo uses the same architecture and the same RL algorithm as AMORTIX, but each training epoch starts from a single carbon atom (one starting
structure, 160 samples), so it never sees a scaffold during training. At test
time it receives the scaffold as its initial graph; since all actions are
additive, the scaffold is preserved by construction.

\begin{table}[t]
    \centering
    \small
    \caption{Experiment-specific hyperparameters for goal-directed scaffold decoration tasks. The AMORTIX variants train until either their maximum budget of 50,000 oracle calls is exhausted or 500 epochs are completed (whichever comes first); GraphXForm runs until the maximum number of epochs is completed.}
    \label{tab:experiment_configs}
    \resizebox{\textwidth}{!}{%
    \begin{tabular}{l p{2.0cm} p{2.0cm} p{2.0cm} p{2.0cm} p{2.0cm}}
        \toprule
         & \textbf{AMORTIX-DeNovo} & \textbf{AMORTIX-REINFORCE} & \textbf{AMORTIX-GRPO} & \textbf{AMORTIX-PPO} & \textbf{GraphXForm} \\
        \midrule
        \textbf{Initialization} & Single Carbon & Scaffolds + C\textsuperscript{*} & Scaffolds + C\textsuperscript{*} & Scaffolds + C\textsuperscript{*} & Single Carbon \\
        \textbf{Method} & Dr. GRPO & REINFORCE\textsuperscript{\dag} & Dr. GRPO & PPO (value critic) & TASAR (SIL) \\
        \textbf{Grouping} & False & False & True & False & False \\
        \textbf{Starting Structures per Epoch} & 1 & 10 & 10 & 10 & 1 \\
        \textbf{Group Size ($G$)\textsuperscript{\ddag}} & 160 & 16 & 16 & 16 & 160 \\
        \textbf{Total Batch Size} & $\approx 160$ & $\approx 160$ ($16 \times 10$) & $\approx 160$ ($16 \times 10$) & $\approx 160$ ($16 \times 10$) & N/A\textsuperscript{\S} \\
        \textbf{Max Epochs} & 500 & 500 & 500 & 500 & 500 \\
        \bottomrule
        \\
        \multicolumn{6}{p{\dimexpr\textwidth-2\tabcolsep\relax}}{\footnotesize \textsuperscript{*}In each epoch, the model is initialized with 1 single carbon atom and 9 scaffolds randomly sampled from the library.} \\
        \multicolumn{6}{l}{\footnotesize \textsuperscript{\dag}Implemented as GRPO without grouping (Global Baseline).} \\
        \multicolumn{6}{p{\dimexpr\textwidth-2\tabcolsep\relax}}{\footnotesize \textsuperscript{\ddag}For the AMORTIX variants, $G$ completions are sampled independently for each of the $B$ starting structures; for GraphXForm, this is the beam width of its search.} \\
        \multicolumn{6}{p{\dimexpr\textwidth-2\tabcolsep\relax}}{\footnotesize \textsuperscript{\S}GraphXForm utilizes elite replay buffer imitation learning; parameter updates are performed on state-action pairs sampled from a cumulative buffer of top molecules rather than an on-policy batch.}
    \end{tabular}}
\end{table}

\textbf{Baseline Note:} The \textbf{GraphXForm} column represents the GraphXForm training methodology~\cite{graphxform}. Unlike our RL-based approach, this baseline is trained via self-improvement learning (SIL) to mimic high-scoring trajectories discovered by the \textit{Take a Step and Reconsider} (TASAR)~\cite{tasar} search heuristic. It serves as a direct comparison between RL and search-based imitation learning. We utilize the default hyperparameters of the GraphXForm implementation, modifying only the beam width and the number of epochs to ensure consistency with our other experiments.

\paragraph{PPO-Specific Settings.}
The AMORTIX-PPO ablation shares the data-sampling configuration of AMORTIX-GRPO above, replacing group-relative advantages with a scalar value-critic head trained jointly on the shared Graph Transformer backbone. Because environmental feedback is strictly terminal, we calculate per-step advantages using undiscounted Monte Carlo returns ($A_t = R_{\text{final}} - V(s_t)$), equivalent to GAE with $\gamma=1.0$ and $\lambda=1.0$. Additional PPO-specific hyperparameters include: clip range = $0.2$, epochs per update = $1$, value loss coefficient = $0.5$, and entropy coefficient = $0.0$.

Advantages are normalized per batch (zero mean, unit variance), as is standard
for PPO, whereas the GRPO and REINFORCE variants use raw advantages.

\paragraph{Prodrug Objective Settings.}
For the \textit{Prodrug} experiments (not listed in Table~\ref{tab:experiment_configs}), the configuration followed the AMORTIX-GRPO setup but utilized 4 specific parent molecules (morphine, GABA, nipecotic acid, and
aspirin) and, additionally, a single carbon atom as starting points, with 32 samples per parent and thus $32 \times 5 = 160$ trajectories per epoch. The policy was initialized from the ChEMBL-pretrained checkpoint and fine-tuned for 50 epochs with full-parameter updates (Adam, learning rate $10^{-4}$, no entropy bonus), using group-relative reward centering within each parent's group. This costs a maximum of $32 \times 5 \times 50 = 8{,}000$ unique reward evaluations.

\subsection{LibINVENT Implementation Details}
\label{app:libinvent}

We used the LibINVENT scaffold-decoration model from REINVENT 4.7.15~\cite{reinvent4}: an encoder--decoder RNN with attention that autoregressively generates the two R-groups for a scaffold annotated with attachment-point markers. Both the prior and the agent were initialized from the publicly released LibINVENT prior (trained on a subset of ChEMBL). Training used REINVENT's DAP (Distribution Augmented Policy) reinforcement-learning strategy.

The key training hyperparameters are:
\begin{itemize}
    \setlength\itemsep{0.1em}
    \item \textbf{General Parameters:} Batch size = 64; Randomize SMILES = True.
    \item \textbf{Learning Strategy:} Type = DAP; reward strength $\sigma = 128$; Optimizer = Adam; Learning rate = $10^{-4}$.
    \item \textbf{Diversity Filter:} Type = IdenticalMurckoScaffold; Bucket size = 25; Min score = 0.4.
    \item \textbf{Training Budget:} RL steps until 50{,}000 oracle calls are consumed, matching AMORTIX's budget.
\end{itemize}

\subsection{DrugEx v3 Implementation Details}
\label{app:drugex}

We used the official DrugEx v3~\cite{drugex_v3} Graph Transformer architecture, initializing the policy from the pretrained \texttt{DrugEx\_PT\_ChEMBL27} checkpoint. Optimization was performed with the \texttt{FragGraphExplorer} module, which fine-tunes the agent against the multi-objective reward via Pareto Crowding Distance sorting, using a frozen copy of the pretrained model as a mutation network.

The key training hyperparameters are:
\begin{itemize}
    \setlength\itemsep{0.1em}
    \item \textbf{General Parameters:} Minibatch size = 64; Exploration rate ($\epsilon$) = 0.1.
    \item \textbf{Training Budget:} Max epochs = 500.
    \item \textbf{Exploration Settings:}
    \begin{itemize}
        \item beta = 0.0 (reward baseline coefficient).
        \item no\_multifrag\_smiles = True (single connected component).
    \end{itemize}
    \item \textbf{Reward Scheme:} Pareto Crowding Distance.
\end{itemize}

\subsection{GenMol Implementation Details}
\label{app:genmol}
We used the official GenMol V2 implementation~\cite{genmol} (\texttt{model\_v2.ckpt}) in its scaffold-decoration mode with the repository's default sampling parameters for this task (\texttt{hparams\_v2.yaml}), including molecular context guidance. Oracle-guided optimization runs for 32 generations. In each generation, GenMol
samples 32 candidates for the scaffold and scores them with the oracle; the scores
guide the next generation, and the best molecule found over all generations is
reported ($\approx$1{,}000 oracle calls per scaffold). Oracle calls are counted only
for molecules that have not been scored before. We note that all
molecules generated by GenMol contained the given input scaffold as a substructure.

The key hyperparameters are:
\begin{itemize}
    \setlength\itemsep{0.1em}
    \item \textbf{Sampling Parameters:} Softmax temperature = 1.5; Randomness = 2; Gamma ($\gamma$) = 0.3 (guidance scale $w=2$).
    \item \textbf{Generation Constraints:} Minimum added length (min\_add\_len) = 24.
    \item \textbf{Attachment Tokens:} Up to 3 attachment tokens ([*]) placed per iteration.
    \item \textbf{Population Settings:} \texttt{NUM\_GENERATIONS} = 32; \texttt{NUM\_SAMPLES} = 32.
\end{itemize}

\subsection{Mol~GA Implementation Details}
\label{app:molga}
Mol~GA is initialized from the given scaffold molecule and run until an oracle budget of 10,000 distinct (canonical-SMILES) molecules is exhausted, or 100 generations are reached. Only offspring that retain the scaffold are scored, and the budget is checked at the start of each generation, with the final scoring batch truncated to the remaining budget. To enforce the scaffold constraint, offspring violating the scaffold substructure requirement are assigned a fitness of zero.
The hyperparameters are listed in Table~\ref{tab:molga_hyperparams}.
\begin{table}[h]
    \centering
    \caption{Hyperparameters and configuration details for Mol~GA.}
    \label{tab:molga_hyperparams}
    \begin{tabular}{l c}
        \toprule
        \textbf{Parameter} & \textbf{Value} \\
        \midrule
        \multicolumn{2}{c}{\textit{General GA Settings}} \\
        Starting Population & 1 (the starting scaffold) \\
        Offspring Size & 1,000 \\
        Max Generations & 100 \\
        Selection Strategy & Greedy (Top-N) \\
        Parent Sampling Strategy & Uniform Quantile Sampling (25 quantiles) \\
        Mating Pool Size & 2,000 \\
        \midrule
        \multicolumn{2}{c}{\textit{Offspring Generation Strategy}} \\
        Crossover Fraction & 0.90 \\
        Mutation-Only Fraction & 0.10 \\
        Post-Crossover Mutation Rate & 0.01 \\
        Crossover Type & 50\% Ring / 50\% Non-Ring \\
        \midrule
        \multicolumn{2}{c}{\textit{Graph Mutation Probabilities}} \\
        Insert Atom & 0.15 \\
        Append Atom & 0.15 \\
        Change Bond Order & 0.14 \\
        Delete Cyclic Bond & 0.14 \\
        Add Ring & 0.14 \\
        Delete Atom & 0.14 \\
        Change Atom Type & 0.14 \\
        \bottomrule
    \end{tabular}
\end{table}

\clearpage

\section{Goal-Directed Scaffold Decoration: Full Results}\label{app:scaffold_full}
\subsection{Training-Algorithm Ablation Across 23 Objectives}
\label{app:ablation_full}
We repurpose the 22 PMO scoring functions, together with Kinase MPO, as
scaffold-decoration rewards on the same held-out test scaffolds, using the same
pretrained checkpoint and the same $\approx$50{,}000-call training budget for all
AMORTIX variants (GRPO, PPO, and REINFORCE). LibINVENT is trained until it consumes the same budget.
Table~\ref{tab:ablation_full} gives per-objective scores; the summary in
Table~\ref{tab:ablation_summary} (main text) is computed from it.

\begin{table}[h]
\caption{Scaffold decoration across 23 objectives. Mean $\pm$ SD over 3 scaffold splits (one training run per split), where each score is the mean objective value over the 500 held-out test scaffolds of a split. The best mean per
objective is in bold. Objectives marked
with $^\dagger$ are the three tasks of Table~\ref{tab:main_performance}.}
\label{tab:ablation_full}
\begin{center}
\begin{small}
\begin{sc}
\resizebox{\textwidth}{!}{
\begin{tabular}{lcccc}
\toprule
Objective & GRPO & PPO & REINFORCE & LibINVENT \\
\midrule
Albuterol Similarity & \textbf{0.584} $\pm$ 0.002 & 0.579 $\pm$ 0.004 & 0.557 $\pm$ 0.018 & 0.259 $\pm$ 0.019 \\
Amlodipine MPO & \textbf{0.462} $\pm$ 0.003 & 0.454 $\pm$ 0.013 & 0.441 $\pm$ 0.093 & 0.313 $\pm$ 0.031 \\
Celecoxib Rediscovery & \textbf{0.397} $\pm$ 0.002 & 0.345 $\pm$ 0.004 & 0.394 $\pm$ 0.024 & 0.232 $\pm$ 0.021 \\
Deco Hop & 0.848 $\pm$ 0.005 & \textbf{0.850} $\pm$ 0.015 & 0.842 $\pm$ 0.092 & 0.535 $\pm$ 0.005 \\
DRD2 & \textbf{0.900} $\pm$ 0.002 & 0.600 $\pm$ 0.003 & 0.540 $\pm$ 0.028 & 0.681 $\pm$ 0.040 \\
Fexofenadine MPO & \textbf{0.802} $\pm$ 0.004 & 0.785 $\pm$ 0.013 & 0.797 $\pm$ 0.100 & 0.609 $\pm$ 0.008 \\
GSK3$\beta^\dagger$ & \textbf{0.775} $\pm$ 0.005 & 0.725 $\pm$ 0.017 & 0.757 $\pm$ 0.084 & 0.414 $\pm$ 0.122 \\
Isomers C7H8N2O2 & 0.317 $\pm$ 0.002 & \textbf{0.328} $\pm$ 0.003 & 0.319 $\pm$ 0.031 & 0.010 $\pm$ 0.004 \\
Isomers C9H10N2O2PF2Cl & 0.407 $\pm$ 0.008 & \textbf{0.416} $\pm$ 0.019 & 0.387 $\pm$ 0.116 & 0.041 $\pm$ 0.004 \\
JNK3$^\dagger$ & \textbf{0.796} $\pm$ 0.004 & 0.728 $\pm$ 0.008 & 0.754 $\pm$ 0.092 & 0.261 $\pm$ 0.065 \\
Kinase MPO$^\dagger$ & \textbf{0.619} $\pm$ 0.004 & 0.552 $\pm$ 0.015 & 0.475 $\pm$ 0.120 & 0.453 $\pm$ 0.007 \\
Median 1 & \textbf{0.212} $\pm$ 0.006 & 0.183 $\pm$ 0.015 & 0.194 $\pm$ 0.058 & 0.061 $\pm$ 0.006 \\
Median 2 & \textbf{0.260} $\pm$ 0.004 & 0.236 $\pm$ 0.013 & 0.237 $\pm$ 0.071 & 0.122 $\pm$ 0.008 \\
Mestranol Similarity & \textbf{0.585} $\pm$ 0.003 & 0.479 $\pm$ 0.009 & 0.484 $\pm$ 0.065 & 0.175 $\pm$ 0.024 \\
Osimertinib MPO & \textbf{0.862} $\pm$ 0.002 & 0.848 $\pm$ 0.005 & 0.836 $\pm$ 0.050 & 0.732 $\pm$ 0.008 \\
Perindopril MPO & \textbf{0.582} $\pm$ 0.007 & 0.445 $\pm$ 0.029 & 0.411 $\pm$ 0.123 & 0.187 $\pm$ 0.010 \\
QED & \textbf{0.854} $\pm$ 0.002 & 0.811 $\pm$ 0.003 & 0.802 $\pm$ 0.033 & 0.690 $\pm$ 0.009 \\
Ranolazine MPO & \textbf{0.826} $\pm$ 0.002 & 0.812 $\pm$ 0.007 & 0.819 $\pm$ 0.054 & 0.591 $\pm$ 0.031 \\
Scaffold Hop & 0.486 $\pm$ 0.006 & \textbf{0.488} $\pm$ 0.019 & 0.478 $\pm$ 0.116 & 0.399 $\pm$ 0.006 \\
Sitagliptin MPO & \textbf{0.292} $\pm$ 0.010 & 0.258 $\pm$ 0.034 & 0.131 $\pm$ 0.039 & 0.019 $\pm$ 0.009 \\
Thiothixene Rediscovery & 0.469 $\pm$ 0.011 & \textbf{0.480} $\pm$ 0.034 & 0.431 $\pm$ 0.129 & 0.241 $\pm$ 0.015 \\
Troglitazone Rediscovery & \textbf{0.445} $\pm$ 0.003 & 0.425 $\pm$ 0.014 & 0.410 $\pm$ 0.081 & 0.158 $\pm$ 0.010 \\
Zaleplon MPO & \textbf{0.335} $\pm$ 0.002 & 0.302 $\pm$ 0.005 & 0.280 $\pm$ 0.039 & 0.178 $\pm$ 0.026 \\
\midrule
Sum & 13.115 & 12.129 & 11.776 & 7.361 \\
Best on (of 23) & 18 & 5 & 0 & 0 \\
\bottomrule
\end{tabular}
}
\end{sc}
\end{small}
\end{center}
\end{table}

\paragraph{Statistical details.}
Across the 23 objectives, GRPO has the higher mean score than REINFORCE on 22
(Wilcoxon signed-rank on paired per-objective means, $p=4.8\times10^{-7}$) and
than PPO on 18 ($p=1.0\times10^{-4}$). The median gain is $+0.026$ over REINFORCE
and $+0.024$ over PPO, and it is concentrated in a minority of objectives, such as
DRD2 ($+0.36$ over REINFORCE) and Perindopril MPO ($+0.17$). PPO's five wins are
all by at most $0.011$. The split SD of REINFORCE exceeds $0.09$ on 9 of 23
objectives (maximum $0.129$), compared with a maximum of $0.011$ for GRPO and
$0.034$ for PPO. Where the mean difference to REINFORCE is small, it often lies
within REINFORCE's own split-to-split spread.

\begin{figure}[h]
    \centering
    \includegraphics[width=\linewidth]{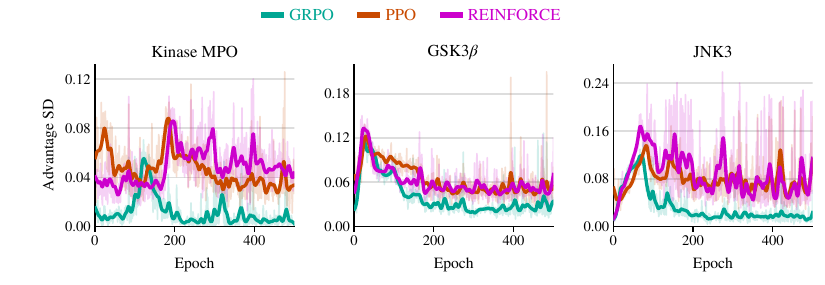}
    \caption{\textbf{Advantage SD during training.} Standard
    deviation of the advantage over the scaffold-started trajectories of each
    batch, before any normalization. Thin lines: per epoch; thick lines: Gaussian-smoothed ($\sigma=3$
epochs). One run per method on a single scaffold
split. After the initial phase of fast reward
    improvement, GRPO gives the smallest spread on all three
    tasks.}
    \label{fig:adv_sd_raw}
\end{figure}

\paragraph{Advantage signal.}
Figure~\ref{fig:adv_sd_raw} shows the unnormalized advantage SD of the runs in
Figure~\ref{fig:variance}. All variants peak during the phase of fast reward
improvement; afterwards GRPO gives the lowest spread on all
three tasks. Figure~\ref{fig:variance} shows the
ratio of the two spreads in the form of one minus its square, which does not
depend on the scale of the rewards; the ratio itself is $1$ for a global
baseline, and for PPO it averages $1.8$, $1.3$ and $2.7$ over the first 20
epochs on Kinase MPO, GSK3$\beta$ and JNK3. Both quantities are computed over the
144 trajectories started from scaffolds in each batch (nine groups of 16; the
single-carbon start is excluded). For PPO the advantage is the reward minus the
critic's value of the starting structure, before the per-batch normalization used
in training. Because each group mean contains the sample itself, centering within
groups of 16 lowers the sample SD by a factor of $\sqrt{15/16}\approx0.97$ even
if scaffolds explained none of the variance; in the variance-removed form of Figure~\ref{fig:variance}, this corresponds to about $0.06$.

\begin{figure}[h]
    \centering
    \includegraphics[width=0.97\linewidth]{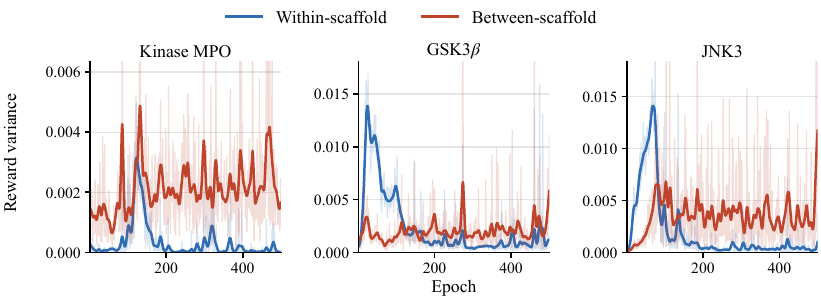}
    \caption{\textbf{Within- and between-scaffold reward variance} for the runs of
    Figure~\ref{fig:variance}. Within: variance among samples of the same scaffold
    (the squared advantage SD); between: variance of the scaffold means. Thin
    lines: per epoch; thick lines: Gaussian-smoothed ($\sigma=3$ epochs). Shows one run
    per task.}
    \label{fig:within_between}
\end{figure}

\paragraph{Within- and between-scaffold variance.}
For group-centered advantages, the variance of the rewards in a batch is exactly
the sum of two parts: the variance among samples of the same scaffold, which
equals the variance of the advantage, and the variance between the scaffold
means. The fraction of the reward variance that centering removes
(Figure~\ref{fig:variance}, bottom) is the between-scaffold part divided by the
total. Figure~\ref{fig:within_between} shows both parts for the same runs. The
early dip in that fraction (Section~\ref{subsec:ablation}) comes from the
within-scaffold part. While the reward is rising, the policy is still changing
and its samples for a single scaffold differ widely; the within-scaffold variance
peaks at roughly ten to twenty times its late level and falls back as the policy
settles, which is when the fraction recovers. The variance between scaffolds stays
at about the same level on average around the peak.

\subsection{Sample Diversity}\label{app:diversity}
For computing molecule diversity metrics, we use Morgan fingerprints (radius 2, 2048 bits) with Tanimoto
distance. IntDiv1 is one minus the mean pairwise Tanimoto similarity of a set of
molecules~\cite{benhenda2017chemgan, polykovskiy2020moses}; we report it among
the completions of the same scaffold, averaged over scaffolds (within), and among
one completion from each of different scaffolds (across). Uniqueness is the
fraction of distinct molecules among the completions of a scaffold, and Murcko
diversity is the number of distinct Bemis--Murcko scaffolds divided by the number
of molecules. \#Circles is the size of the largest subset of molecules whose pairwise Tanimoto
distances all exceed $t=0.75$, the threshold recommended by Xie et
al.~\cite{xie2023circles}, computed on the pooled completions. The diversity results are reported in Table~\ref{tab:diversity}.

\begin{table}[h]
\caption{Diversity of the generated molecules.
The last block averages 23 objectives. IntDiv1 (W): within the completions of a scaffold; (A): across scaffolds.}
\label{tab:diversity}
\begin{center}
\begin{small}
\begin{sc}
\resizebox{\textwidth}{!}{
\begin{tabular}{llccccc}
\toprule
Task & Method & IntDiv1 (W) & IntDiv1 (A) & Uniq. & Murcko & \#Circles \\
\midrule
GSK3$\beta$ & GRPO & 0.491 & 0.781 & 0.762 & 0.682 & 87 \\
 & PPO & 0.505 & 0.761 & 0.751 & 0.680 & 113 \\
 & REINFORCE & 0.550 & 0.784 & 0.940 & 0.627 & 111 \\
\midrule
JNK3 & GRPO & 0.459 & 0.764 & 0.687 & 0.598 & 91 \\
 & PPO & 0.528 & 0.765 & 0.741 & 0.673 & 118 \\
 & REINFORCE & 0.540 & 0.778 & 0.879 & 0.590 & 114 \\
\midrule
Kinase MPO & GRPO & 0.353 & 0.678 & 0.908 & 0.735 & 68 \\
 & PPO & 0.310 & 0.684 & 0.881 & 0.686 & 64 \\
 & REINFORCE & 0.135 & 0.668 & 0.195 & 0.186 & 7 \\
\midrule
Mean of 23 & GRPO & 0.387 & 0.733 & 0.546 & 0.291 & 102.6 \\
 & PPO & 0.405 & 0.735 & 0.534 & 0.289 & 100.3 \\
 & REINFORCE & 0.418 & 0.743 & 0.619 & 0.283 & 104.7 \\
\bottomrule
\end{tabular}
}
\end{sc}
\end{small}
\end{center}
\end{table}

\subsection{Kinase MPO Sub-Task Breakdown}\label{app:mpo_breakdown}

\begin{table}[H]
    \caption{Detailed Kinase MPO sub-task breakdown. Values represent the global mean $\pm$ standard deviation across 3 scaffold splits. \textbf{AMORTIX} denotes the default GRPO variant. While instance optimizers default to the easier heuristic metric QED, AMORTIX successfully prioritizes the complex biological targets under hard topological constraints while keeping synthetic accessibility in a comparable range.}
    \label{tab:mpo_breakdown_appendix}
    \begin{center}
        \begin{small} 
        \begin{sc}       
        \setlength{\tabcolsep}{4pt} 
        \resizebox{\textwidth}{!}{
            \begin{tabular}{lccccc}
                \toprule
                Method & Obj. Score & GSK3$\beta$ ($\uparrow$) & JNK3 ($\uparrow$) & QED ($\uparrow$) & SA ($\downarrow$) \\
                \midrule
                \multicolumn{6}{l}{\textit{Amortized Policies}} \\
                LibINVENT & $0.453 \pm 0.007$ & $0.280 \pm 0.012$ & $0.164 \pm 0.019$ & $0.596 \pm 0.016$ & $3.052 \pm 0.045$ \\
                GraphXForm & $0.409 \pm 0.003$ & $0.064 \pm 0.007$ & $0.024 \pm 0.005$ & $0.793 \pm 0.008$ & $3.205 \pm 0.023$ \\
                DrugEx v3  & $0.354 \pm 0.003$ & $0.036 \pm 0.009$ & $0.018 \pm 0.003$ & $0.613 \pm 0.004$ & $3.266 \pm 0.056$ \\[1ex]
                AMORTIX-DeNovo & $0.397 \pm 0.005$ & $0.071 \pm 0.012$ & $0.030 \pm 0.004$ & $0.710 \pm 0.014$ & $3.007 \pm 0.059$ \\
                AMORTIX-REINFORCE & $0.475 \pm 0.120$ & $0.352 \pm 0.293$ & $0.285 \pm 0.205$ & $0.505 \pm 0.223$ & $3.178 \pm 2.068$ \\
                AMORTIX-PPO & $0.552 \pm 0.015$ & $0.606 \pm 0.011$ & $0.478 \pm 0.039$ & $0.380 \pm 0.025$ & $3.304 \pm 0.329$ \\
                \textbf{AMORTIX} & \textbf{0.619 $\pm$ 0.004} & \textbf{0.743 $\pm$ 0.007} & \textbf{0.588 $\pm$ 0.011} & $0.381 \pm 0.004$ & $3.114 \pm 0.075$ \\
                \midrule
                \multicolumn{6}{l}{\textit{Instance Optimizers}} \\
                Mol~GA & $0.461 \pm 0.013$ & $0.236 \pm 0.025$ & $0.127 \pm 0.027$ & $0.813 \pm 0.020$ & $4.006 \pm 0.278$ \\
                GenMol & $0.479 \pm 0.012$ & $0.188 \pm 0.018$ & $0.061 \pm 0.027$ & \textbf{0.880 $\pm$ 0.035} & \textbf{2.929 $\pm$ 0.046} \\
                \bottomrule
            \end{tabular}
            }
        \end{sc}
        \end{small}
    \end{center}
\end{table}

Relative to Mol~GA (the instance optimizer with stronger results on the two biological targets), the Kinase MPO reward of AMORTIX gains $0.127$ from GSK3$\beta$,
$0.115$ from JNK3 and $0.025$ from SA$'$, and loses $0.108$ through QED, a net
gain of $0.159$. The QED sacrifice is thus substantial, but far smaller than the
gain on the biological targets.

\paragraph{Success Rate.} On Kinase MPO, 17.8\% $\pm$ 9.3\% of AMORTIX's molecules meet a conservative
success criterion on all four sub-objectives (GSK3$\beta \geq 0.5$, JNK3 $\geq
0.5$, QED $\geq 0.6$, SA $\leq 4$), against 6.8\% $\pm$ 12.6\% for PPO and 9.1\%
$\pm$ 15.1\% for REINFORCE (mean $\pm$ SD over three splits).

\section{Prodrug Transfer: Full Results}
\label{app:prodrug}

\begin{figure}[h]
  \centering
  \includegraphics[width=\linewidth]{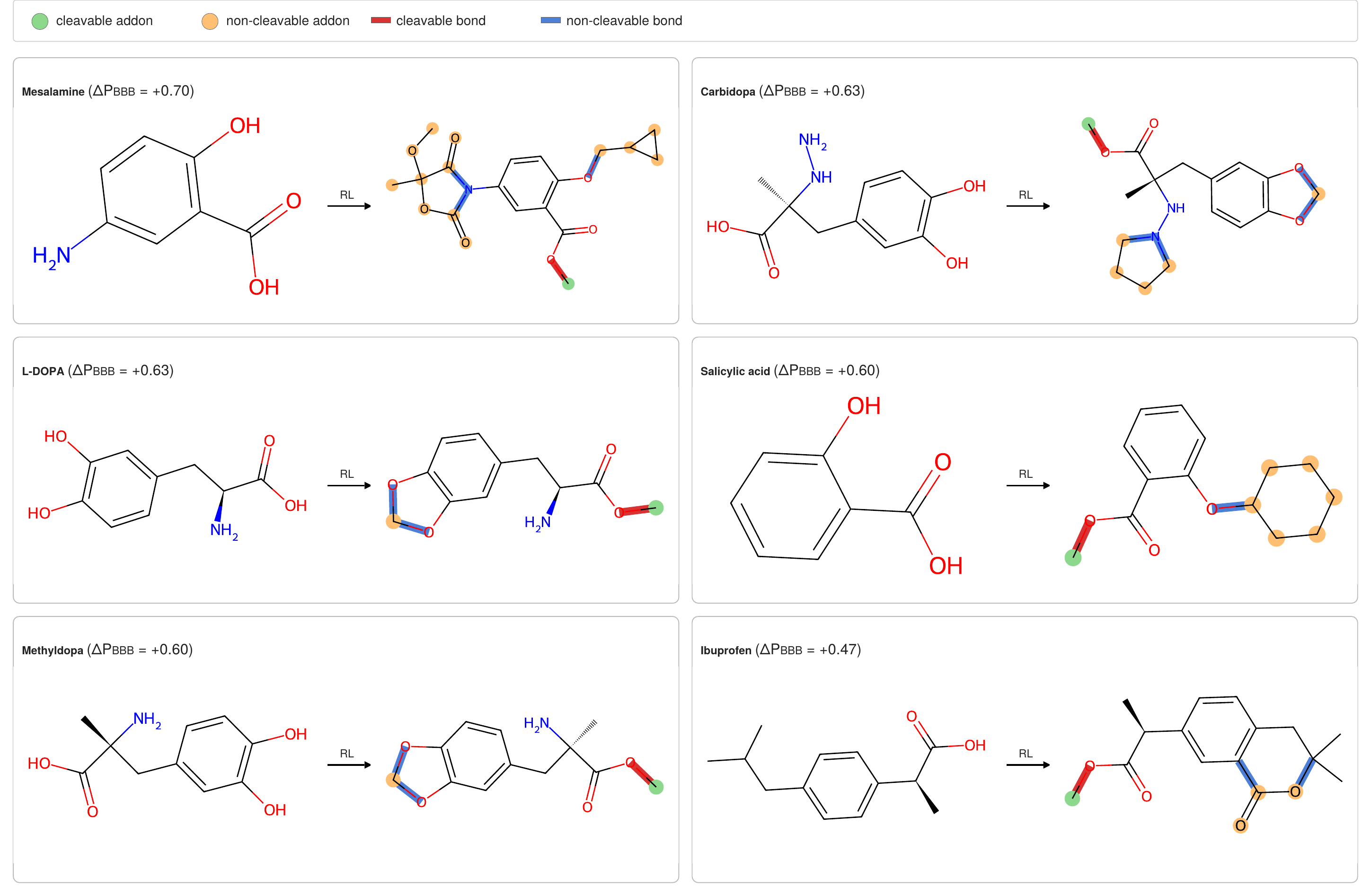}
  \caption{Top six partial prodrugs by $\Delta P_{\text{BBB}}$: each has a cleavable methyl ester (green addon, red bond) and other modifications that are not cleavable (orange addons, blue bonds).}
  \label{fig:prodrug_partials}
\end{figure}

\begin{figure}[H]
  \centering
  \includegraphics[width=\linewidth]{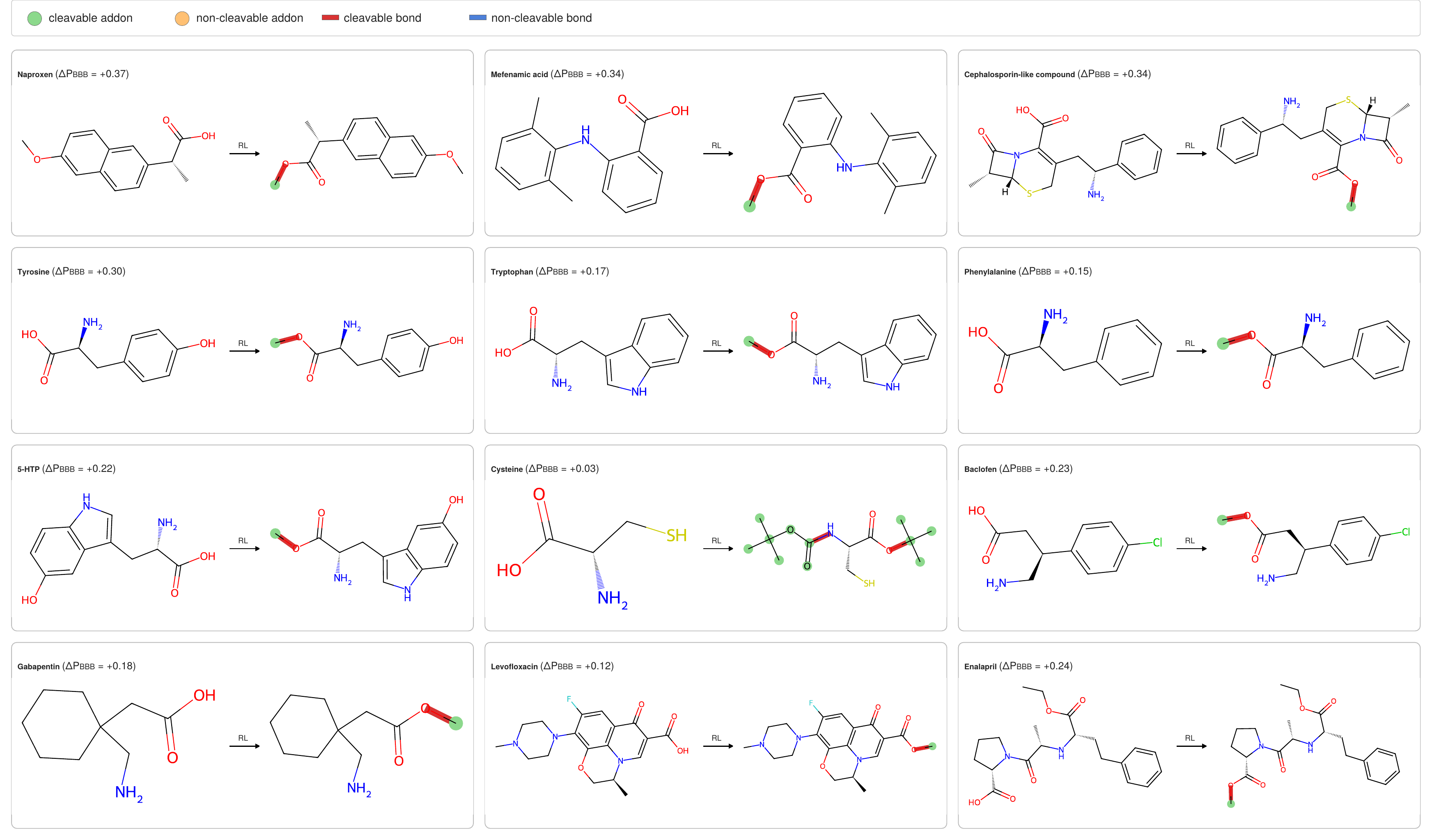}
  \caption{Pure prodrugs (continued from Figure~\ref{fig:prodrug_main}).
  The remaining 12 of the 16 pure prodrugs, each shown next to its parent.
  Eleven are methyl esters of a carboxylic acid. Cysteine carries a carbamate
  and an ester, and its thiol is left unmodified.}
  \label{fig:prodrug_other_pures}
\end{figure}

\paragraph{Reward distribution.}
The reward is deliberately strict: a candidate scores only if it passes the
drug-likeness (QED) threshold and contains at least one cleavable junction. Under this
criterion, 36.3\% of the 1{,}664 evaluation candidates, more than one in three,
score above zero, with a median of $0.31$, a mean of $0.37$ and a best of
$0.95$ among them. The remaining 63.7\% receive zero reward, as the
multiplicative form implies. When considering the best of the 32 candidates drawn per parent drug, the mean
reward is $0.46$ across the 52 parents, and for 33 of them the best candidate carries at least one cleavable handle.

\paragraph{Parent drugs.}
Tables~\ref{tab:prodrug_train_parents} and~\ref{tab:prodrug_test_parents} list the
four exemplar parents used for fine-tuning and the 52 held-out parents used for
evaluation, with the SMILES strings as used in the experiments.

\begin{table}[h]
\caption{Exemplar parent drugs used for fine-tuning.}
\label{tab:prodrug_train_parents}
\begin{center}
\footnotesize
\setlength{\tabcolsep}{4pt}
\begin{tabular}{L{3.0cm} L{9.0cm}}
\toprule
Parent drug & SMILES \\
\midrule
Morphine & \texttt{\seqsplit{CN1CC[C@]23[C@@H]4[C@H]1CC5=C2C(=C(C=C5)O)O[C@H]3[C@H](C=C4)O}} \\
GABA & \texttt{\seqsplit{C(CC(=O)O)CN}} \\
Nipecotic acid & \texttt{\seqsplit{C1CNCCC1C(=O)O}} \\
Aspirin & \texttt{\seqsplit{CC(=O)OC1=CC=CC=C1C(=O)O}} \\
\bottomrule
\end{tabular}
\end{center}
\end{table}

\begingroup
\setlength{\tabcolsep}{4pt}
\begin{longtable}{>{\footnotesize}r >{\footnotesize}L{3.0cm} >{\footnotesize}L{8.6cm}}
\caption{The 52 held-out parent molecules used for evaluation, grouped by compound class.}
\label{tab:prodrug_test_parents}\\
\toprule
\textbf{No.} & \textbf{Parent drug} & \textbf{SMILES} \\
\midrule
\endfirsthead
\multicolumn{3}{l}{\footnotesize\textit{Table~\thetable{} (continued)}}\\
\toprule
\textbf{No.} & \textbf{Parent drug} & \textbf{SMILES} \\
\midrule
\endhead
\bottomrule
\endfoot
\multicolumn{3}{l}{\footnotesize\textit{Catecholamines and amino acids}} \\
1 & Dopamine & \texttt{\seqsplit{C1=CC(=C(C=C1CCN)O)O}} \\
2 & Norepinephrine & \texttt{\seqsplit{C1=CC(=C(C=C1C(CN)O)O)O}} \\
3 & Epinephrine & \texttt{\seqsplit{CNC[C@@H](C1=CC(=C(C=C1)O)O)O}} \\
4 & L-DOPA & \texttt{\seqsplit{C1=CC(=C(C=C1C[C@@H](C(=O)O)N)O)O}} \\
5 & Carbidopa & \texttt{\seqsplit{C[C@@](Cc1ccc(O)c(O)c1)(NN)C(=O)O}} \\
6 & 5-HTP & \texttt{\seqsplit{C1=CC2=C(C=C1O)C(=CN2)C[C@@H](C(=O)O)N}} \\
7 & Tyrosine & \texttt{\seqsplit{N[C@@H](Cc1ccc(O)cc1)C(=O)O}} \\
8 & Tryptophan & \texttt{\seqsplit{N[C@@H](Cc1c[nH]c2ccccc12)C(=O)O}} \\
9 & Phenylalanine & \texttt{\seqsplit{N[C@@H](Cc1ccccc1)C(=O)O}} \\
10 & Histidine & \texttt{\seqsplit{N[C@@H](Cc1c[nH]cn1)C(=O)O}} \\
\midrule
\multicolumn{3}{l}{\footnotesize\textit{Small amino acids and GABA analogues}} \\
11 & Glycine & \texttt{\seqsplit{NCC(=O)O}} \\
12 & Taurine & \texttt{\seqsplit{NCCS(=O)(=O)O}} \\
13 & $\beta$-Alanine & \texttt{\seqsplit{NCCC(=O)O}} \\
14 & Vigabatrin & \texttt{\seqsplit{C=CC(N)CCC(=O)O}} \\
15 & Pregabalin & \texttt{\seqsplit{CC(C)C[C@@H](CN)CC(=O)O}} \\
16 & Gabapentin & \texttt{\seqsplit{NCC1(CC(=O)O)CCCCC1}} \\
17 & Baclofen & \texttt{\seqsplit{NC[C@@H](CC(=O)O)c1ccc(Cl)cc1}} \\
18 & Tranexamic acid & \texttt{\seqsplit{NCC1CCC(C(=O)O)CC1}} \\
\midrule
\multicolumn{3}{l}{\footnotesize\textit{Carboxylic acid NSAIDs and analgesics}} \\
19 & Ibuprofen & \texttt{\seqsplit{CC(C)Cc1ccc(cc1)[C@H](C)C(=O)O}} \\
20 & Naproxen & \texttt{\seqsplit{COc1ccc2cc([C@@H](C)C(=O)O)ccc2c1}} \\
21 & Ketoprofen & \texttt{\seqsplit{OC(=O)C(C)c1cccc(c1)C(=O)c1ccccc1}} \\
22 & Diclofenac & \texttt{\seqsplit{OC(=O)Cc1ccccc1Nc1c(Cl)cccc1Cl}} \\
23 & Indomethacin & \texttt{\seqsplit{Cc1c(CC(=O)O)c2cc(OC)ccc2n1C(=O)c1ccc(Cl)cc1}} \\
24 & Salicylic acid & \texttt{\seqsplit{OC(=O)c1ccccc1O}} \\
25 & Mefenamic acid & \texttt{\seqsplit{Cc1cccc(C)c1Nc1ccccc1C(=O)O}} \\
\midrule
\multicolumn{3}{l}{\footnotesize\textit{Diuretics and cardiovascular drugs}} \\
26 & Bumetanide & \texttt{\seqsplit{CCCCNc1cc(C(=O)O)c(S(N)(=O)=O)cc1Oc1ccccc1}} \\
27 & Furosemide & \texttt{\seqsplit{OC(=O)c1cc(NCc2ccco2)c(S(N)(=O)=O)cc1Cl}} \\
28 & Probenecid & \texttt{\seqsplit{CCCN(CCC)S(=O)(=O)c1ccc(C(=O)O)cc1}} \\
29 & Methyldopa & \texttt{\seqsplit{C[C@@](N)(Cc1ccc(O)c(O)c1)C(=O)O}} \\
\midrule
\multicolumn{3}{l}{\footnotesize\textit{Antibiotics and antivirals}} \\
30 & Penicillin G & \texttt{\seqsplit{CC1(C)S[C@@H]2[C@H](NC(=O)Cc3ccccc3)C(=O)N2[C@H]1C(=O)O}} \\
31 & Ampicillin & \texttt{\seqsplit{CC1(C)S[C@@H]2[C@H](NC(=O)[C@@H](N)c3ccccc3)C(=O)N2[C@H]1C(=O)O}} \\
32 & Cephalosporin-like compound & \texttt{\seqsplit{C[C@H]1[C@H]2SCC(=C(N2C1=O)C(=O)O)C[C@@H](N)c1ccccc1}} \\
33 & Ciprofloxacin & \texttt{\seqsplit{O=C(O)c1cn(C2CC2)c2cc(N3CCNCC3)c(F)cc2c1=O}} \\
34 & Levofloxacin & \texttt{\seqsplit{C[C@H]1COc2c(N3CCN(C)CC3)c(F)cc3c(=O)c(C(=O)O)cn1c23}} \\
35 & Acyclovir & \texttt{\seqsplit{Nc1nc2c(ncn2COCCO)c(=O)[nH]1}} \\
36 & Zidovudine (AZT) & \texttt{\seqsplit{Cc1cn([C@H]2C[C@H](N=[N+]=[N-])[C@@H](CO)O2)c(=O)[nH]c1=O}} \\
37 & Lamivudine & \texttt{\seqsplit{Nc1ccn([C@@H]2CS[C@H](CO)O2)c(=O)n1}} \\
\midrule
\multicolumn{3}{l}{\footnotesize\textit{Anticancer agents}} \\
38 & Methotrexate & \texttt{\seqsplit{CN(Cc1cnc2nc(N)nc(N)c2n1)c1ccc(C(=O)N[C@@H](CCC(=O)O)C(=O)O)cc1}} \\
39 & Doxorubicin & \texttt{\seqsplit{C[C@H]1O[C@H](C[C@H](N)[C@@H]1O)O[C@H]1C[C@@](O)(C(=O)CO)Cc2c(O)c3c(=O)c4cccc(OC)c4c(=O)c3c(O)c12}} \\
40 & Crizotinib & \texttt{\seqsplit{C[C@H](Oc1cc(-c2cnn(C3CCNCC3)c2)cnc1N)c1c(Cl)ccc(F)c1Cl}} \\
41 & Imatinib & \texttt{\seqsplit{Cc1ccc(NC(=O)c2ccc(CN3CCN(C)CC3)cc2)cc1Nc1nccc(-c2cccnc2)n1}} \\
\midrule
\multicolumn{3}{l}{\footnotesize\textit{Endogenous and metabolic compounds}} \\
42 & Glutamic acid & \texttt{\seqsplit{N[C@@H](CCC(=O)O)C(=O)O}} \\
43 & Aspartic acid & \texttt{\seqsplit{N[C@@H](CC(=O)O)C(=O)O}} \\
44 & Cysteine & \texttt{\seqsplit{N[C@@H](CS)C(=O)O}} \\
45 & N-acetylcysteine & \texttt{\seqsplit{CC(=O)N[C@@H](CS)C(=O)O}} \\
46 & Glutathione & \texttt{\seqsplit{N[C@@H](CCC(=O)N[C@@H](CS)C(=O)NCC(=O)O)C(=O)O}} \\
47 & Ascorbic acid & \texttt{\seqsplit{OC[C@H](O)[C@H]1OC(=O)C(O)=C1O}} \\
48 & Niacin & \texttt{\seqsplit{OC(=O)c1cccnc1}} \\
49 & Pyridoxine & \texttt{\seqsplit{Cc1ncc(CO)c(CO)c1O}} \\
\midrule
\multicolumn{3}{l}{\footnotesize\textit{Other}} \\
50 & Mesalamine & \texttt{\seqsplit{Nc1ccc(O)c(C(=O)O)c1}} \\
51 & Captopril & \texttt{\seqsplit{CC(CS)C(=O)N1CCC[C@H]1C(=O)O}} \\
52 & Enalapril & \texttt{\seqsplit{CCOC(=O)[C@H](CCc1ccccc1)N[C@@H](C)C(=O)N1CCC[C@H]1C(=O)O}} \\
\end{longtable}
\endgroup

\end{document}